\documentclass{article} % For LaTeX2e
\usepackage{spirit_arxiv,times}

\usepackage{amsmath,amsfonts,bm}

\def\eqref#1{equation~\ref{#1}}
\def\1{\bm{1}}

\DeclareMathAlphabet{\mathsfit}{\encodingdefault}{\sfdefault}{m}{sl}
\SetMathAlphabet{\mathsfit}{bold}{\encodingdefault}{\sfdefault}{bx}{n}

\usepackage[utf8]{inputenc} % allow utf-8 input
\usepackage[T1]{fontenc}    % use 8-bit T1 fonts
\usepackage{hyperref}       % hyperlinks
\usepackage{url}            % simple URL typesetting
\usepackage{booktabs}       % professional-quality tables
\usepackage{amsfonts}       % blackboard math symbols
\usepackage{nicefrac}       % compact symbols for 1/2, etc.
\usepackage{microtype}      % microtypography
\usepackage{xcolor}         % colors
\usepackage{graphicx} 
\usepackage{amsmath}
\usepackage{subcaption}
\usepackage{algorithm}
\usepackage{algorithmic}

\usepackage{wrapfig}

\usepackage{wrapfig}
\title{SPIRIT: Sparse Physics-Informed Inverse Rendering for SAR Imaging}

\author{
Yimeng Liu$^{1}$ \quad
Maolin Gan$^{1}$ \quad
Jingkai Lin$^{1}$ \quad
Jialuo Du$^{1}$ \quad
Mi Zhang$^{2}$ \quad
Zhichao Cao$^{1}$
\\[1.2ex]
$^{1}$ Michigan State University
\quad
$^{2}$ Ohio State University
}

\begin{document}

\maketitle

\begin{abstract}
Radio-frequency (RF) and wave-based imaging systems play a critical role in sensing under adverse conditions, but are fundamentally constrained by a trade-off between collection speed and spatial resolution.
Reducing measurements to enable rapid collection of fragment geometric information across sensing dimensions, leading to highly non-injective inverse problems that destabilize conventional image-domain reconstruction and super-resolution methods.
Synthetic aperture radar (SAR) exemplifies this challenge in sparse-aperture sampling.
We introduce SPIRIT, a physics-informed inverse rendering framework for sparse-aperture SAR.
Rather than enhancing formed images, SPIRIT reconstructs an explicit three-dimensional surface.
By adopting an optimization strategy and explicitly modeling visibility, phase coherence, and wave propagation, SPIRIT aggregates distributed geometric features across sparse aperture views into a coherent reconstruction.
Extensive experiments on synthetic and real-world data demonstrate that SPIRIT recovers accurate geometry and signal under extreme sparsity, reducing data collection by over 90\% while achieving low geometric and signal error and consistently outperforming baselines.
\end{abstract}

\section{Introduction}

High-resolution radio-frequency (RF) imaging fundamentally depends on large effective apertures. In practice, physically large arrays are often impractical, and modern systems instead rely on sensor motion to synthesize virtual apertures, as in synthetic aperture radar (SAR) \cite{Cumming2005SAR,soumekh1999synthetic}. While effective, high-fidelity SAR reconstruction typically requires dense spatial sampling, which can lead to long data-collection times even for relatively small objects \cite{Yanik2020Development}. Reducing measurements is therefore attractive for fast sensing, but under severe aperture undersampling the degradation is not merely a loss of image sharpness. Sparse-aperture SAR becomes a coherent inverse problem in which large portions of the spatial-frequency domain are unobserved, producing interference, sidelobes, and geometric ambiguity \cite{Patel2010Compressed,potter2010sparsity}. In this regime, the task is better formulated as sparse-aperture inverse imaging: infer a scene whose coherent forward response explains the retained measurements while regularizing components that are not uniquely determined by the sparse aperture.

Existing image super-resolution paradigms are poorly matched to this setting. Image-domain methods recover plausible high-frequency content from learned priors \cite{dong2015image,liang2021swinir,duan2025dit4sr}, but they do not explicitly enforce the geometry-dependent phase relationships of coherent RF sensing. Recent neural radar methods instead optimize implicit or explicit scene representations through differentiable radar models \cite{radarfields2024,geraf2025,zhang2025spinrv2,kung2025radarsplat}. These systems establish the value of physics-aware radar reconstruction, but their acquisition geometries and reconstruction targets differ from the extreme scan-line sparsity studied here. When most aperture support is removed, the forward operator admits a large null space: several geometrically distinct scenes can be compatible with the retained measurements.

Sparse coherent RF reconstruction is difficult for two coupled reasons. First, severe aperture sparsity creates non-injectivity because missing angular and spatial-frequency support cannot be recovered from the measurements alone \cite{potter2010sparsity,Li2008MIMO}. Second, phase-sensitive optimization is highly non-convex: at millimeter-wave frequencies, sub-millimeter geometric perturbations can induce large phase changes, so direct complex-valued fitting can converge to measurement-consistent but geometrically incorrect solutions \cite{Antun2020Instabilities,Arridge2019Solving}. Analytic renderers introduce a third practical issue: real measurements contain hardware response, material-dependent scattering, calibration error, and multipath that are only approximately represented by tractable forward models.

\begin{wrapfigure}{r}{0.48\linewidth}
  \centering
  \includegraphics[width=\linewidth]{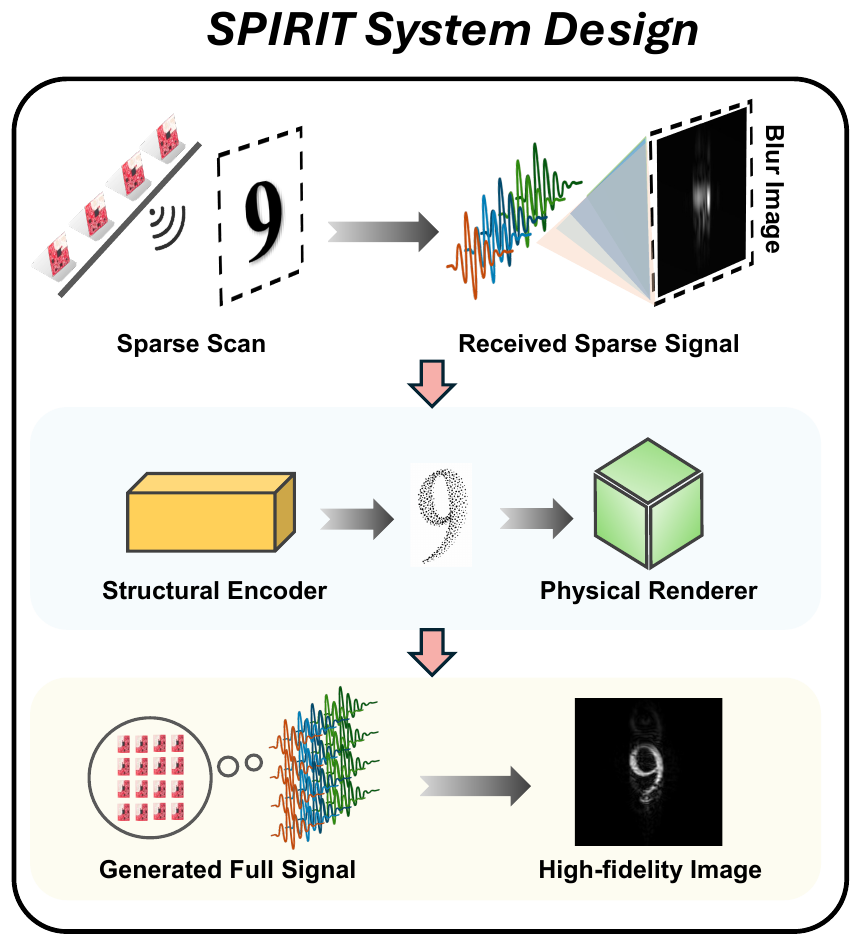}
\caption{SPIRIT formulates sparse-aperture SAR reconstruction as physics-informed inverse rendering.}
  \label{fig:teaser}
\end{wrapfigure}

We introduce \textbf{SPIRIT}, a sparse physics-informed inverse-rendering framework for SAR imaging. Rather than enhancing a formed image, SPIRIT reconstructs an explicit three-dimensional surface whose forward-rendered measurements are constrained by the observed sparse signals. The framework combines a structured surface representation, a differentiable single-scattering RF renderer with visibility and phase modeling, and a two-stage optimization schedule. Stage~I establishes a geometry-stable hypothesis using objectives that are less sensitive to phase wrapping. Stage~II then introduces complex-valued consistency together with a constrained residual pathway for bounded model mismatch.

Our formulation targets severe scan-line sparsity, where only a small subset of aperture positions is retained. The missing support makes the inverse problem fundamentally non-unique; retained views provide complementary evidence but do not preserve all information available from the full aperture. SPIRIT therefore should be interpreted as selecting a structured, measurement-consistent solution under explicit geometric and physical inductive biases, not as recovering absent information without assumptions.

\paragraph{Contributions.}
Our contributions are threefold: (i) we formulate extreme sparse-aperture SAR as a structure-centric inverse-rendering problem that makes non-injectivity and prior dependence explicit; (ii) we introduce a geometry-first reconstruction architecture that combines a structured surface prior, differentiable RF rendering, and staged phase-sensitive refinement; and (iii) through object-disjoint controlled experiments, we show that measurement fit and geometric correctness can decouple under severe sparsity which motivates separating structural hypothesis formation from phase-sensitive refinement. We further evaluate the final system on an object-disjoint MITO test split in Appendix~\ref{sec:real_world}. Stage~I geometry learning uses synthetic measurements, while a disjoint real training split is used only for Stage~II residual adaptation; the resulting experiment tests physical measurements with hardware, calibration, material, and multipath effects not reproduced exactly by the synthetic pipeline. Because the full pipeline uses real adaptation, we do not describe this experiment as zero-shot synthetic-to-real transfer.

\section{Related Work}
\label{sec:related}

Sparse-aperture reconstruction lies at the intersection of image super-resolution, neural RF scene reconstruction, and classical sparse inverse problems. The important distinction is not simply model family, but what is observed, what latent scene representation is optimized, and which ambiguities are resolved by data versus priors.

\begin{table*}[t]
\centering
\caption{Scope comparison of representative reconstruction paradigms. Entries summarize the published input domain, scene representation, forward model, and primary reconstruction regime; the table is descriptive rather than a performance ranking.}
\label{tab:related_comparison}
\scriptsize
\setlength{\tabcolsep}{2.4pt}
\begin{tabular}{p{2.15cm}p{2.15cm}p{2.45cm}p{2.85cm}p{3.05cm}}
\toprule
\textbf{Method} & \textbf{Input} & \textbf{Representation} & \textbf{Forward / Prior Model} & \textbf{Primary Target / Regime} \\
\midrule
SwinIR~\cite{liang2021swinir} & formed image & pixel/Transformer features & learned image prior & image restoration / super-resolution \\
DiT4SR~\cite{duan2025dit4sr} & formed image & diffusion Transformer & learned generative prior & real-world image super-resolution \\
\midrule
Radar Fields~\cite{radarfields2024} & raw FMCW frequency data & implicit neural field & physics-informed FMCW sensor model & radar scene reconstruction from measured trajectories \\
GeRaF~\cite{geraf2025} & RF measurements & neural SDF / reflectance field & physics-based RF volumetric rendering & near-range 3D geometry reconstruction \\
SpINRv2~\cite{zhang2025spinrv2} & complex FMCW spectrum & implicit neural representation & differentiable frequency-domain synthesis & volumetric FMCW reconstruction \\
RadarSplat~\cite{kung2025radarsplat} & radar images + poses & Gaussian primitives & radar rendering + noise/multipath model & driving-scene reconstruction and synthesis \\
Back-Projection Diffusion~\cite{li2024backprojection} & wideband scattered measurements & conditional diffusion in BP latent & wave-based back-projection structure & wideband inverse scattering, including sub-Nyquist settings \\
\midrule
\textbf{SPIRIT (Ours)} & sparse FMCW phase history + poses & structured explicit surface & differentiable single-scattering RF renderer & isolated-object geometry under extreme scan-line sparsity \\
\bottomrule
\end{tabular}
\end{table*}

\subsection{Image Super-Resolution}
Image super-resolution methods learn mappings from degraded or low-resolution images to visually sharper outputs. CNN/Transformer regressors such as SRCNN and SwinIR \cite{dong2015image,liang2021swinir} and recent diffusion-based systems \cite{duan2025dit4sr,clearsr2024} can synthesize plausible high-frequency detail. Sparse SAR differs because aperture removal changes the sensing operator itself: the task is to infer a scene consistent with complex-valued coherent measurements, not merely to restore an already formed image. We therefore use image-domain super-resolution as a conceptual contrast rather than a direct solution to the problem studied here.

\subsection{Neural RF Inverse Rendering}
Physics-aware radar reconstruction increasingly combines neural scene representations with differentiable forward models. Radar Fields learns implicit geometry and reflectance directly from raw FMCW measurements \cite{radarfields2024}; GeRaF reconstructs geometry using neural implicit representations and RF volumetric rendering \cite{geraf2025}; SpINRv2 supervises an implicit representation in the complex frequency domain \cite{zhang2025spinrv2}; and RadarSplat uses Gaussian primitives with explicit radar-noise modeling for driving scenes \cite{kung2025radarsplat}. Differentiable SAR rendering similarly enables gradient-based reconstruction from simulated SAR imagery \cite{Huang2022DiffRender}. These works establish that RF reconstruction benefits from sensor-aware forward models, but they use different representations, acquisition support, and scene regimes. Our focus is narrower: isolated-object reconstruction when scan-line support is deliberately reduced to an extreme degree.

\subsection{Sparse Aperture and Inverse Scattering}
Sparse SAR/ISAR has long been studied through compressed sensing and structured regularization \cite{Patel2010Compressed,potter2010sparsity,bi2023mixednorm}. Learning-based approaches include unfolded sparse-aperture imaging and autofocusing \cite{afampnet2021}, learned TomoSAR inversion \cite{qian2021gnet}, regularization-by-denoising for sparse 3D SAR \cite{wang2024redsar}, and self-supervised sparse ISAR reconstruction \cite{wang2025isarnet}. Wideband Back-Projection Diffusion further combines wave-based back-projection structure with a learned diffusion prior for inverse scattering \cite{li2024backprojection}. SPIRIT differs in where the prior is imposed: the latent variable is an explicit surface with deterministic neighborhood structure, and measurement fitting is mediated by a differentiable coherent renderer. Our experiments test whether this geometry-first inductive bias is effective in the controlled, predominantly connected-object regime; we do not claim that a fixed-topology surface is universally preferable to implicit fields, Gaussian primitives, surfels, or hybrid representations. Accordingly, the quantitative baselines in Section~\ref{sec:experiments} are matched controls under a common sparse-aperture protocol rather than claimed faithful reproductions of every cited radar system.

\begin{figure*}[!t]
  \centering
  \includegraphics[width=0.85\linewidth]{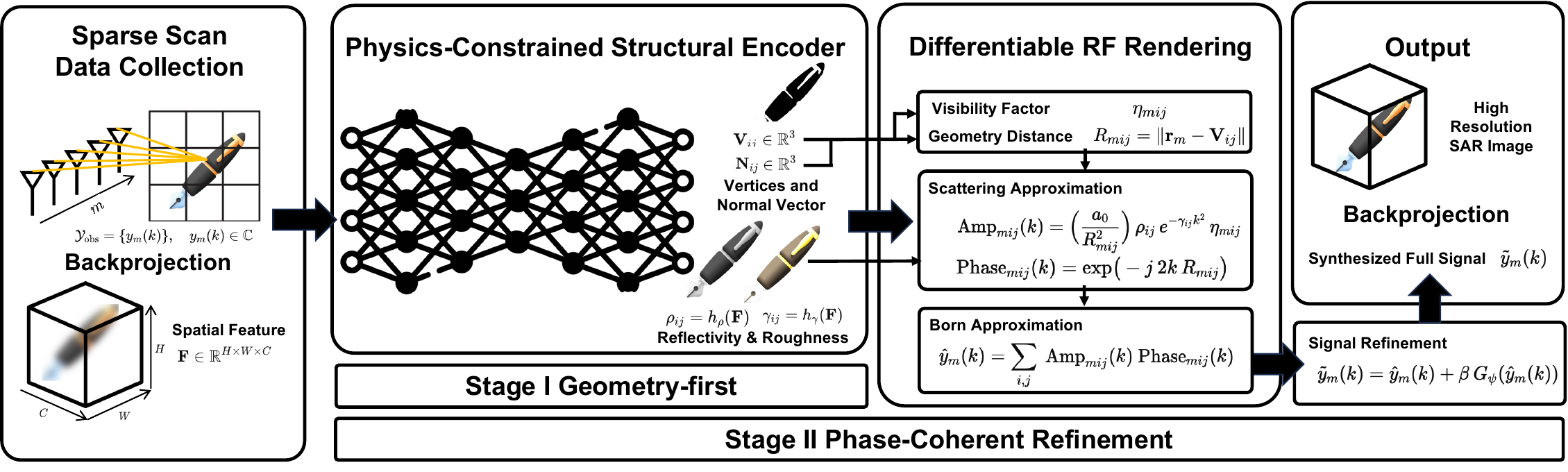}
\caption{
Overview of SPIRIT. Sparse coherent measurements are encoded into an explicit surface representation and evaluated through a differentiable RF forward model. The renderer produces predicted measurements at observed and unobserved aperture locations; consistency with the observed sparse measurements constrains the recovered geometry, while the learned structural prior regularizes unobserved components of the scene.
}

  \label{fig:testbed_overview}
\end{figure*}

\section{Problem Formulation}
\label{sec:formulation}

We formulate sparse-aperture RF reconstruction as a physics-constrained inverse-imaging problem. Let $\mathbf{Y}_{\mathrm{obs}}$ denote coherent measurements retained at a sparse subset of aperture positions. Because this sampling removes portions of the spatial-frequency and angular support, the corresponding forward operator is non-injective: multiple scene configurations may explain the same sparse observations.
Our objective is to seek a physically plausible scene representation whose forward response is consistent with the observed measurements while being regularized by explicit structural priors. Because the operator is non-injective, this objective does not identify a unique ground-truth scene from the sparse measurements alone; it selects one measurement-consistent hypothesis under the representation and regularization assumptions.

\subsection{Structure-Centric Inverse Imaging}

Conventional super-resolution methods cast reconstruction as a pixel-domain mapping
$f: I_{\mathrm{LR}} \rightarrow I_{\mathrm{HR}}$.
In RF imaging, however, apparent resolution loss arises from limited aperture coverage, leaving large portions of the spectrum unobserved under sparse scanning, as detailed in Appendix~\ref{app:AA_limitation}.

As detailed in Appendices~\ref{app:AA_phase} and~\ref{app:AA_sar_forward}, RF signals encode scene geometry primarily through the phase accumulated along propagation paths.
When aperture coverage is sparse, the forward operator admits a non-trivial null space, and multiple geometrically distinct scenes may explain the same or nearly indistinguishable sparse measurements.
In this regime, the missing components are not uniquely determined by formed-image intensities alone; any reconstruction of them necessarily relies on additional structural, physical, or learned assumptions.

We therefore solve the sparse-aperture inverse problem in an explicit, structure-centric scene representation.
\begin{equation}
    \mathbf{Y} \approx \mathcal{F}(\mathcal{S}),
\end{equation}
where $\mathcal{F}$ denotes a physics-based forward operator modeling coherent wave propagation, whose derivation is provided in Appendix \ref{app:AA_em_model}.
The scene representation encodes explicit \emph{surface-centric physical structure} rather than image pixels:
\begin{equation}
    \mathcal{S} = \{\mathbf{P}, \mathbf{N}, \boldsymbol{\rho}, \boldsymbol{\gamma}\}.
\end{equation}
Here, $\mathbf{P} \in \mathbb{R}^{N \times 3}$ denotes a structured set of surface points in 3D space.
Surface normals $\mathbf{N} \in \mathbb{R}^{N \times 3}$ are deterministically derived from $\mathbf{P}$ and encode local surface orientation.
Material properties are modeled using two complementary parameters:
$\boldsymbol{\rho} \in \mathbb{R}^N$ represents surface reflectivity, capturing radar cross-section magnitude. 
In contrast, $\boldsymbol{\gamma} \in \mathbb{R}^N$ encodes surface roughness and coherence loss due to sub-wavelength micro-geometry.

\subsection{Physics-Constrained Forward Model}

The mapping from scene structure $\mathcal{S}$ to received signal $\mathbf{Y}$ is governed by electromagnetic wave propagation.
Starting from Maxwell’s equations under time-harmonic excitation, the scattered field satisfies the Helmholtz equation, detailed in Appendix~\ref{app:AA_em_model}.
Specifically, we adopt a Born-style single-scattering surface model, implemented in a Physical-Optics-like form as a discrete sum over surface elements.

We model the measurement process using a differentiable physics-based forward operator $\mathcal{F}_{\mathrm{PO}}$:
\begin{equation}
\begin{aligned}
\hat{y}(\mathbf{r}_m, k)
&=
\mathcal{F}_{\mathrm{PO}}(\mathcal{S}) \\
&\approx
\sum_{i=1}^{N}
\mathcal{V}(\mathbf{x}_i, \mathbf{r}_m)\,
A(\mathbf{x}_i)\,
\boldsymbol{\sigma}_i\,
e^{-j 2 k \|\mathbf{r}_m - \mathbf{x}_i\|},
\end{aligned}
\end{equation}
where $\mathbf{r}_m$ denotes the radar platform position at aperture index $m$, $k$ is the wavenumber, and $\mathbf{x}_i \in \mathbf{P}$ is the $i$-th surface point.
The term $\mathcal{V}(\mathbf{x}_i, \mathbf{r}_m)$ is a differentiable visibility function modeling occlusion, and $A(\mathbf{x}_i)$ accounts for antenna gain and distance-dependent path loss.
The complex scattering coefficient $\boldsymbol{\sigma}_i$ is parameterized by surface reflectivity $\boldsymbol{\rho}_i \in \mathbb{R}$ and roughness $\boldsymbol{\gamma}_i \in \mathbb{R}$.
The approximation arises from a Born-style single-scattering assumption with a Physical-Optics-like surface discretization, under which higher-order multipath and material-dependent effects are neglected for tractability, as detailed in Appendix~\ref{app:AA_em_model} and Appendix~\ref{app:AA_limitation}. Throughout this paper, ``physics-constrained'' means consistency with this explicit approximate forward model; it does not imply exact electromagnetic simulation of all hardware, material, polarization, or multipath effects.

\subsection{Inverse Objective}

Recovering $\mathcal{S}$ from sparse signals is posed as a constrained energy minimization problem:
\begin{equation}
    \label{eq:inverse_objective}
    \min_{\mathcal{S}}
    \;
    \mathcal{L}_{\mathrm{data}}
    \big(
        \mathcal{F}_{\mathrm{PO}}(\mathcal{S}),
        \mathbf{Y}_{\mathrm{obs}}
    \big)
    +
    \lambda\,
    \mathcal{L}_{\mathrm{struct}}(\mathcal{S}),
\end{equation}
where $\mathcal{L}_{\mathrm{data}}$ measures complex-valued discrepancy in both amplitude and phase, and
$\mathcal{L}_{\mathrm{struct}}$ enforces geometric regularity such as surface smoothness and normal consistency \cite{gropp2020implicit}.
Equation~\eqref{eq:inverse_objective} describes the scene-level inverse objective conceptually. In SPIRIT, this inverse map is amortized by the structural encoder: network parameters are learned across the training corpus, while the predicted surface for each input is evaluated through the same forward-consistency and structural terms.

\section{SPIRIT Design}
\label{sec:method}

\subsection{Overview}
SPIRIT reconstructs an explicit scene surface from sparse coherent measurements and uses that surface to predict the corresponding SAR response through a structure-first, physics-constrained inverse-imaging pipeline.
% SPIRIT infers an explicit surface representation whose wave-consistent forward simulation explains the observed sparse signals.
% This design is motivated by the fact that sparse-aperture SAR induces a large null space and a non-injective geometry-to-signal mapping, detailed in Appendix \ref{app:AA_limitation}, making image-domain enhancement fundamentally unstable.
The method consists of three tightly coupled components.
First, a \emph{physics-aligned structural encoder} maps sparse measurements into an explicit surface-based scene representation by injecting known collection geometry and physical constraints.
Second, a \emph{differentiable RF forward renderer} enforces phase-coherent wave propagation, enabling signal-domain supervision to constrain geometry directly.
Third, a \emph{two-stage optimization strategy with signal refinement} stabilizes learning under extreme phase sensitivity and compensates for inevitable model--reality mismatch in analytic RF rendering.
The design addresses two coupled sources of difficulty: sparse aperture sampling leaves a non-trivial null space, while coherent phase makes optimization highly sensitive to small geometric errors. The structured surface prior restricts the ambiguous solution space, and staged optimization delays phase-sensitive fitting until a stable geometric initialization has been established.

% All radar and SAR fundamentals required to interpret these components—including wave propagation, FMCW signal structure, SAR forward modeling, phase sensitivity, and sparse-aperture non-injectivity—are provided in Appendix~A.
% Operator-level derivations and optimization-specific analysis corresponding directly to this section are provided in Appendix~B.

\subsection{Physics-Constrained Structural Encoder}
SPIRIT addresses sparse-aperture non-injectivity by reconstructing an explicit surface rather than a pixel grid, volumetric density field, or unconstrained point set.
Under sparse apertures, high-capacity representations can fit measurement ambiguity without uniquely fixing geometry. We therefore use a structured surface prior with stable local neighborhoods and a compact hypothesis class. This choice is an inductive bias for the isolated, predominantly connected objects studied here rather than a claim of universal superiority over volumetric, implicit, or Gaussian representations.
% Together, these properties reduce the effective degrees of freedom available to exploit the null space induced by sparse apertures,
% while enforcing explicit topology with stable local neighborhoods and strong geometric priors.

Given sparse complex frequency-modulated continuous-wave (FMCW) SAR signals $\mathbf{Y}_{\mathrm{obs}}$ and the corresponding platform poses $\{\mathbf{r}_m\}$ (background in Appendix~\ref{app:AA_fmcw}), SPIRIT first applies a fixed differentiable backprojection operator to expose coarse spatial support:
\begin{equation}
    \mathbf{F} = \mathcal{B}(\mathbf{Y}_{\mathrm{obs}};\{\mathbf{r}_m\}).
\end{equation}
Backprojection maps the signal domain onto a spatial grid, detailed in Appendix \ref{app:AB_dbe}.
On a predefined two-dimensional reference plane, the operator evaluates how strongly a scatterer at each grid location $(x,y)$ would contribute to the received echoes under the known propagation distances and platform geometry; Algorithm~\ref{alg:dbp_forward} gives the discrete implementation.
The output of $\mathcal{B}$ is a feature tensor $\mathbf{F} \in \mathbb{R}^{B \times 3 \times H \times W}$, whose channels store the accumulated real part, imaginary part, and magnitude of the backprojected signal, forming a flattened, defocused spatial energy map.

From the backprojected feature map $\mathbf{F}$, SPIRIT applies a convolutional encoder--decoder to extract structured geometric features from the defocused spatial representation.
We adopt a convolutional encoder–decoder with skip connections,
whose downsampling path aggregates global object extent under heavy sidelobe corruption, while the upsampling path restores spatial alignment.
The network outputs a dense feature map $\mathbf{Z} \in \mathbb{R}^{B \times C \times H \times W}$, where each grid location encodes a local surface descriptor that will subsequently be lifted into an explicit three-dimensional vertex.

To lift the two-dimensional feature grid into an explicit three-dimensional surface, SPIRIT employs a structured grid deformation strategy.
Rather than predicting an unordered point cloud, we interpret the $H \times W$ feature grid as a deformable surface mesh with fixed topology.
% This design ensures that neighboring grid cells in the feature map correspond to neighboring vertices in three-dimensional space, yielding spatial continuity and stable normal estimation.
To prevent degenerate solutions under sparse scans, we enforce bounded support via a global--local structure.
A global head predicts a coarse axis-aligned bounding box $(\mathbf{v}_{\min}, \mathbf{v}_{\max}) \in \mathbb{R}^3 \times \mathbb{R}^3$ that captures the overall spatial extent of the object.
Conditioned on the dense feature map $\mathbf{Z}$, a local head predicts normalized coordinates $\boldsymbol{\delta}_{i,j} \in [0,1]^3$ for each grid cell.
Final vertex positions are then obtained by
\begin{equation}
    \mathbf{V}_{i,j} = \mathbf{v}_{\min} + \boldsymbol{\delta}_{i,j}\odot(\mathbf{v}_{\max}-\mathbf{v}_{\min}),
\end{equation}
which constrains all vertices to the predicted object support and reduces unbounded geometric drift under the sparse-aperture null space.
The resulting inductive bias, implementation details, and representation limitations are discussed in Appendix~\ref{app:AB_grid} and Appendix~\ref{app:discussion}.

Surface normals are computed deterministically from the grid topology.
Local tangents are obtained via finite differences along the grid axes, and per-vertex normals are computed by cross-products and normalization:
\begin{equation}
    \mathbf{N}_{i,j} =
    \frac{(\mathbf{V}_{i+1,j}-\mathbf{V}_{i-1,j})\times(\mathbf{V}_{i,j+1}-\mathbf{V}_{i,j-1})}
    {\|(\mathbf{V}_{i+1,j}-\mathbf{V}_{i-1,j})\times(\mathbf{V}_{i,j+1}-\mathbf{V}_{i,j-1})\|_2+\epsilon}.
\end{equation}
This construction enforces local geometric consistency and stable orientation.
In addition, SPIRIT predicts per-vertex reflectivity $\rho_{i,j}$ and roughness $\gamma_{i,j}$, allowing residual signal variation to be explained without forcing geometry to absorb all mismatch.
Normal computation, gradient flow, and regularization are described in Appendix \ref{app:AB_normals}.

\subsection{Differentiable RF Rendering and Signal Refinement}
SPIRIT enforces physical consistency by synthesizing complex-valued radar signals from the predicted surface and directly comparing them to the observed sparse signals.
The renderer implements a differentiable approximation of the SAR forward operator under a single-bounce scattering model grounded in wave propagation and phase accumulation, shown in Appendix \ref{app:AB_renderer}.
For platform position $\mathbf{r}_m$ and wavenumber $k$, the synthesized return is
\begin{equation}
\begin{aligned}
    \hat{y}_{m}(k) =&
    \sum_{i,j}
    \alpha_{m,i,j}(k)\,
    \sigma_{i,j}(k)\,
    \eta_{m,i,j}\, \\
    & \cdot \exp\!\big(-j\,2k\,\|\mathbf{r}_m-\mathbf{V}_{i,j}\|\big),
\end{aligned}
\end{equation}
where $\alpha_{m,i,j}(k)$ captures distance-dependent attenuation and known system scaling, and $\sigma_{i,j}(k)$ is the complex scattering coefficient parameterized by the predicted reflectivity $\rho_{i,j}$ and roughness $\gamma_{i,j}$.
The visibility term $\eta_{m,i,j}\in[0,1]$ is computed via soft backface culling:
\begin{equation}
\begin{aligned}
    \eta_{m,i,j} &= \sigma\!\big(\kappa\,\langle \mathbf{N}_{i,j}, \mathbf{d}_{m,i,j}\rangle\big),\\
    \qquad
    \mathbf{d}_{m,i,j} &= \frac{\mathbf{r}_m-\mathbf{V}_{i,j}}{\|\mathbf{r}_m-\mathbf{V}_{i,j}\|_2},
\end{aligned}
\end{equation}
where $\sigma$ denotes the sigmoid.
Soft visibility provides continuous gradients with respect to surface orientation, whereas a hard visibility decision would introduce zero-gradient discontinuities during optimization.
Renderer gradients and implementation details are given in Appendix~\ref{app:AB_renderer}.

Even with accurate geometry, analytic RF rendering inevitably deviates from real measurements
due to unmodeled effects such as antenna patterns, hardware response, material-dependent scattering, and weak multipath interactions.
This model--reality mismatch, analyzed in Appendix~\ref{app:AA_limitation}, can bias geometric inference if strict signal consistency is enforced too early or too rigidly.

To compensate for these discrepancies without compromising physical interpretability, SPIRIT introduces a signal refinement layer formulated strictly as a residual correction, rather than a primary generative model:
\begin{equation}
\tilde{\mathbf{Y}} = \hat{\mathbf{Y}} + \beta\,\mathcal{G}_{\psi}(\hat{\mathbf{Y}}).
\end{equation}
Here, $\mathcal{G}_\psi$ is a lightweight residual network applied only to the analytically rendered signal, and $\beta$ controls the magnitude of the residual correction. We initialize $\beta$ near zero and regularize the residual toward the identity mapping so that the analytic renderer remains the dominant source of the predicted signal.
This design restricts, but does not mathematically eliminate, non-physical corrections. The residual branch does not itself parameterize scene geometry, but the joint Stage~II objective can still couple signal refinement and geometry updates. The refiner may therefore absorb signal components not represented by the analytic renderer, so signal improvement is interpreted jointly with geometric accuracy rather than as independent evidence of geometric correctness.
The residual formulation and identity regularization are therefore intended to keep corrections small relative to the analytic prediction, but they do not constitute a hard physical guarantee. Architecture and regularization details are provided in Appendix~\ref{app:AB_refiner}.

\subsection{Two-Stage Optimization Strategy}
% Directly optimizing phase-coherent signal losses under sparse apertures is highly non-convex due to phase wrapping and gradient reversal induced by $e^{-j2kR}$ shown in \ref{app:AA_phase}, and further ill-conditioned by sparse-aperture null spaces and grating lobes shown in \ref{app:AA_limitation}.
% SPIRIT stabilizes training using a two-stage optimization strategy that separates geometry bootstrapping from phase-coherent alignment. 

\paragraph{Stage I: Geometry-first bootstrapping.}
We first optimize the encoder--decoder to recover globally consistent geometry and bounded support while avoiding phase-induced local minima.
Stage~I emphasizes objectives that constrain coarse structure but are less sensitive to phase wrapping, including image-domain losses on backprojected reconstructions, magnitude-based signal terms, and geometric regularizers on surface distribution and normals.
The signal refinement layer is turned off at this stage to prevent the model from explaining missing diversity by learning arbitrary signal corrections.
% This stage is explicitly motivated by the phase-sensitivity analysis in Appendix \ref{app:AA_phase} and by the non-injectivity of sparse apertures in Appendix \ref{app:AA_limitation}.

\paragraph{Stage II: Phase-coherent alignment with controlled refinement.}
After Stage I establishes a geometry-stable initialization, Stage II introduces complex-valued signal consistency and activates the residual refinement module. In the final SPIRIT configuration, the geometry backbone remains trainable but is updated with a substantially reduced learning rate together with geometric regularization, while the material parameters and residual refiner are jointly optimized. 
This controlled joint optimization differs from unconstrained signal fitting. Among the tested variants, full signal-driven unfreezing without the refiner substantially degrades geometry, whereas the final joint configuration preserves the Stage~I geometric solution while improving formed-image and signal metrics (Table~\ref{tab:ablation_core}).
Loss composition and training schedules are detailed in Appendix \ref{app:AB_optimization}.

% Together, these design choices directly address the three challenges identified in the introduction:
% bounded surface inference and physics-aligned encoding mitigate sparse-aperture non-injectivity,
% Two-stage optimization stabilizes phase-coherent fitting under extreme geometric sensitivity,
% and signal refinement compensates for inevitable mismatches in the analytic model while preserving physical interpretability.

\begin{figure*}[t]
\centering
\setlength{\tabcolsep}{1pt}

\begin{tabular}{cccccccc}
\textbf{Snacks} & \textbf{Dishtowel} & \textbf{Shoes} & \textbf{Hair Kit} &
\textbf{Pencil Case} & \textbf{Card} & \textbf{Flats} & \textbf{Marker} \\

 % & \rotatebox{90}{GT} 
 % & \rotatebox{90}{Pred} 
 % & \rotatebox{90}{SAR} 
 % & \rotatebox{90}{Pred} \\

% ---------- Row 1: GT Point Cloud ----------
% \rotatebox{90}{GT PC} &
\includegraphics[width=0.115\linewidth]{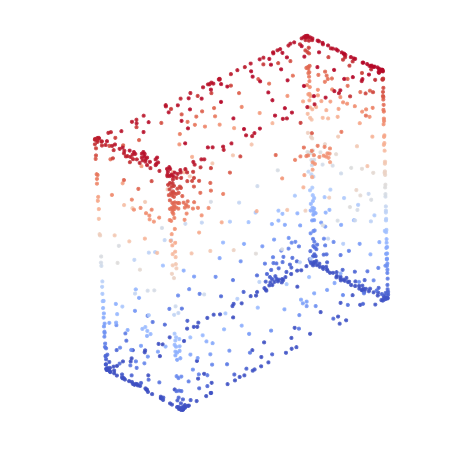} &
\includegraphics[width=0.115\linewidth]{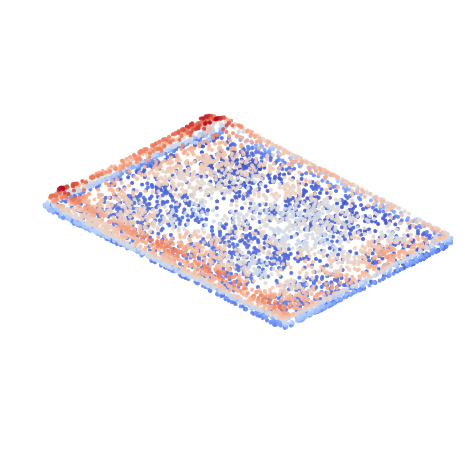} &
\includegraphics[width=0.115\linewidth]{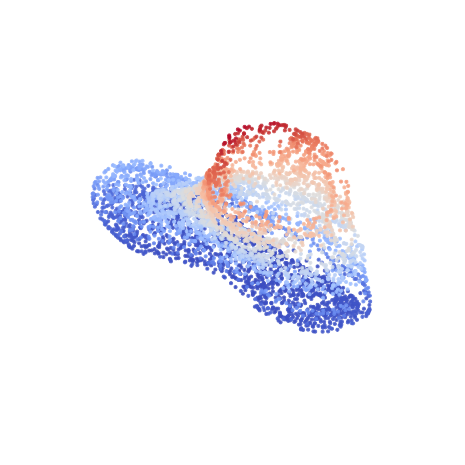} &
\includegraphics[width=0.115\linewidth]{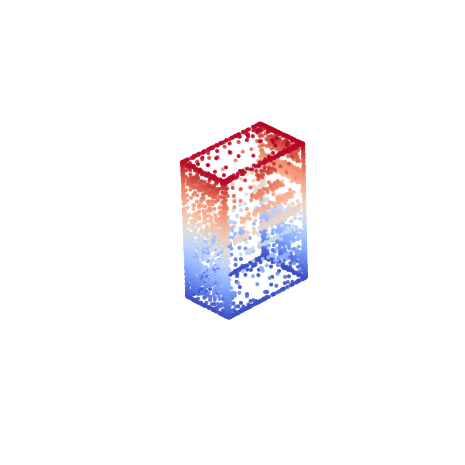} &
\includegraphics[width=0.115\linewidth]{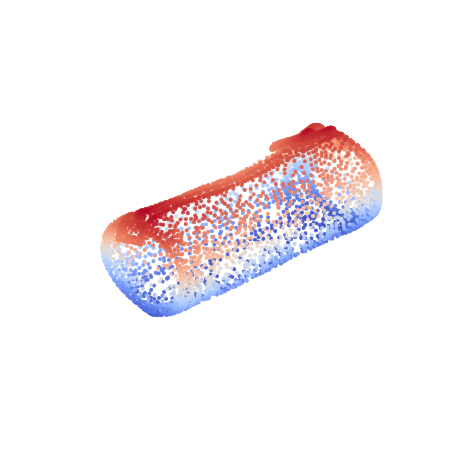} &
\includegraphics[width=0.115\linewidth]{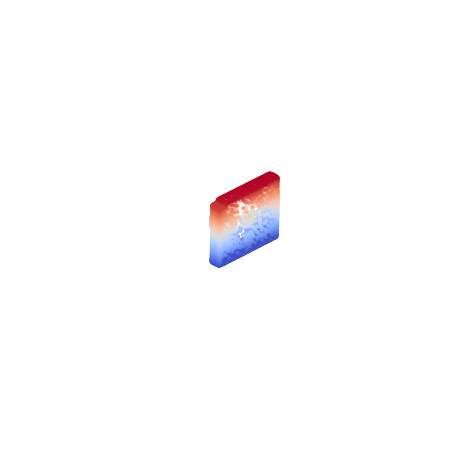} &
\includegraphics[width=0.115\linewidth]{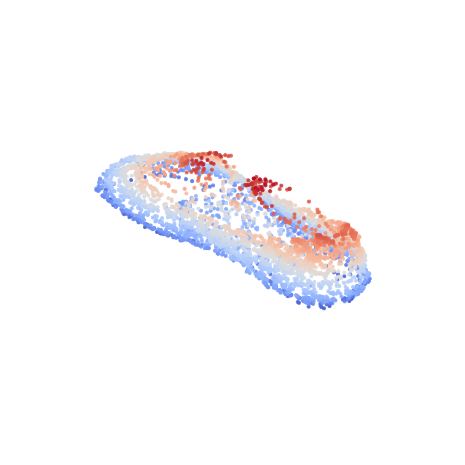} &
\includegraphics[width=0.115\linewidth]{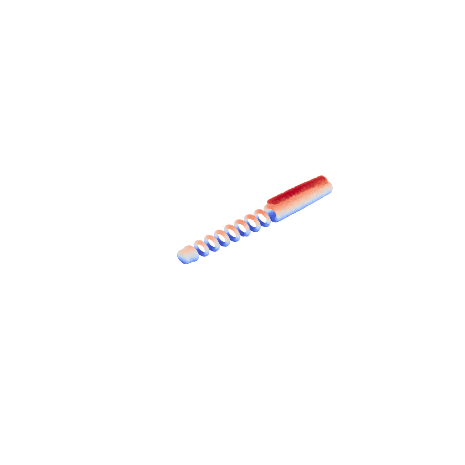} \\[-2pt]

% ---------- Row 2: Pred Point Cloud ----------
% \rotatebox{90}{Pred PC} &
\includegraphics[width=0.115\linewidth]{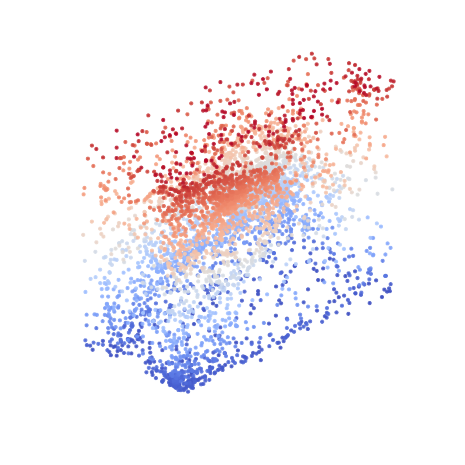} &
\includegraphics[width=0.115\linewidth]{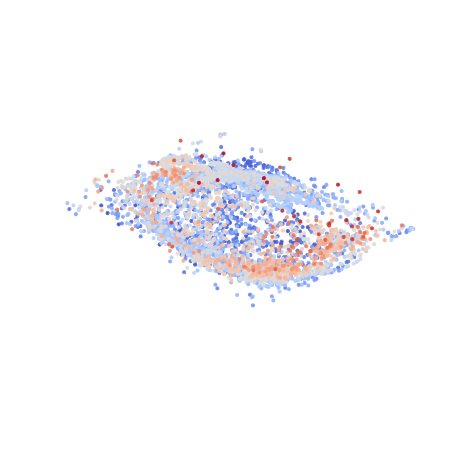} &
\includegraphics[width=0.115\linewidth]{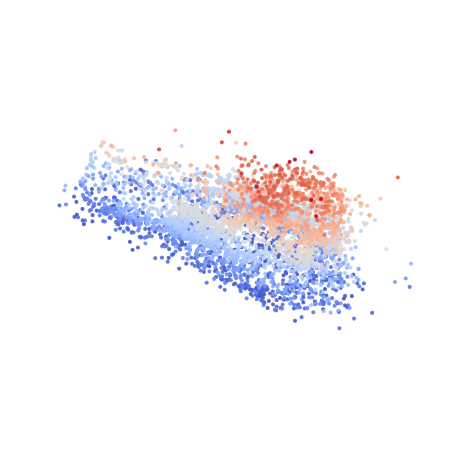} &
\includegraphics[width=0.115\linewidth]{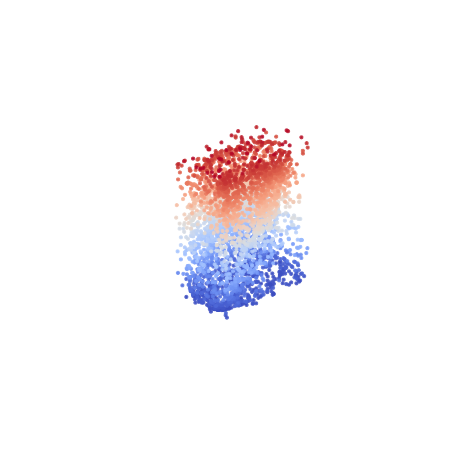} &
\includegraphics[width=0.115\linewidth]{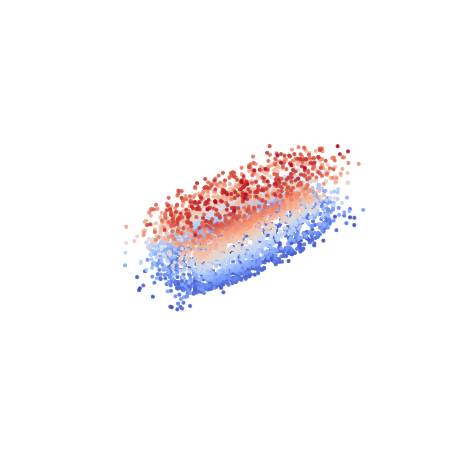} &
\includegraphics[width=0.115\linewidth]{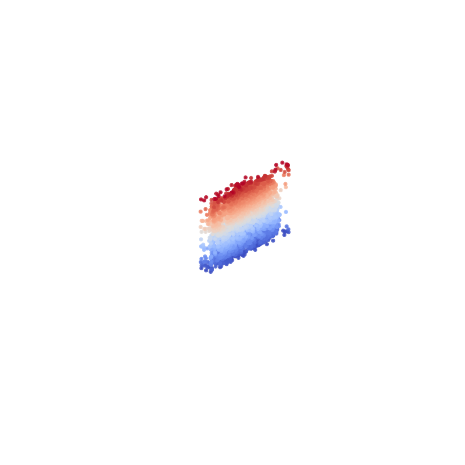} &
\includegraphics[width=0.115\linewidth]{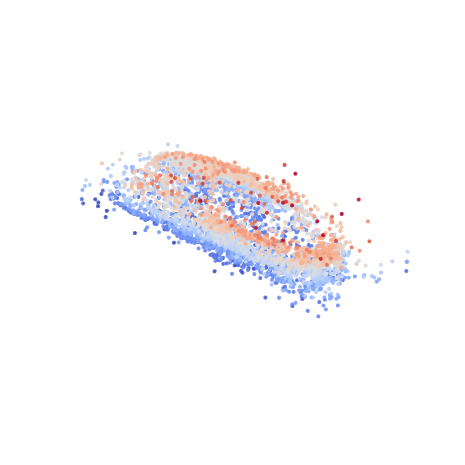} &
\includegraphics[width=0.115\linewidth]{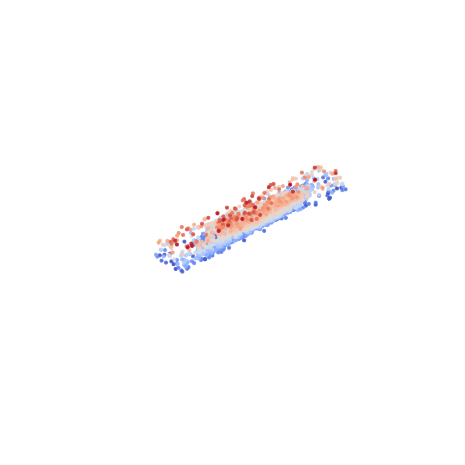} \\[-2pt]

% ---------- Row 3: SAR / Simulation ----------
% \rotatebox{90}{SAR} &
\includegraphics[width=0.115\linewidth]{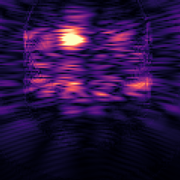} &
\includegraphics[width=0.115\linewidth]{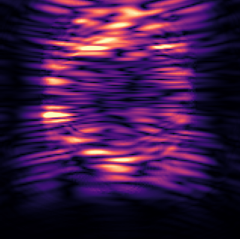} &
\includegraphics[width=0.115\linewidth]{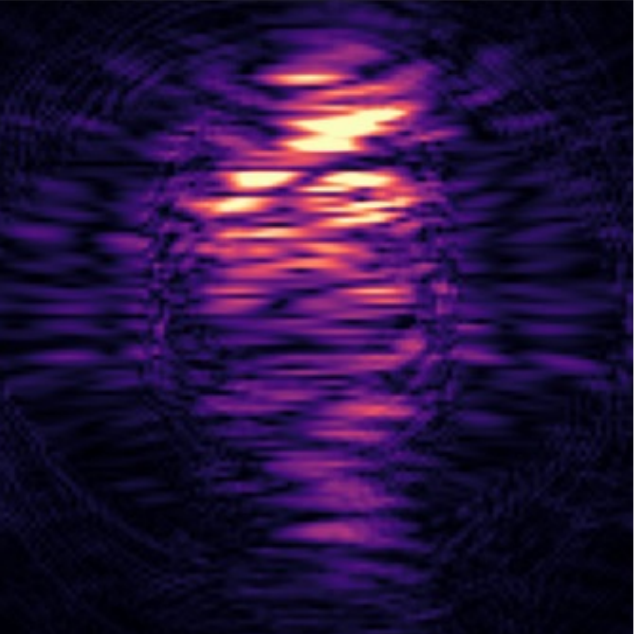} &
\includegraphics[width=0.115\linewidth]{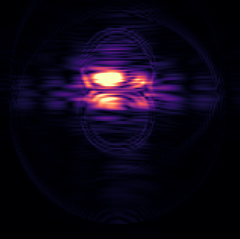} &
\includegraphics[width=0.115\linewidth]{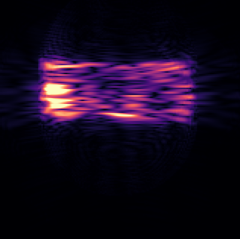} &
\includegraphics[width=0.115\linewidth]{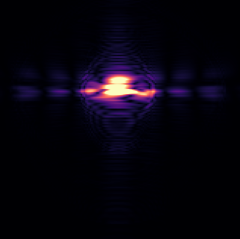} &
\includegraphics[width=0.115\linewidth]{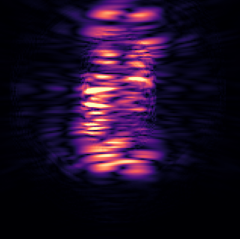} &
\includegraphics[width=0.115\linewidth]{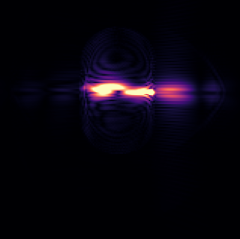} \\[-2pt]

% ---------- Row 4: Pred Signal / Render ----------
% \rotatebox{90}{Pred} &
\includegraphics[width=0.115\linewidth]{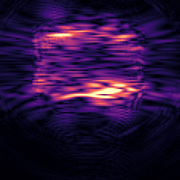} &
\includegraphics[width=0.115\linewidth]{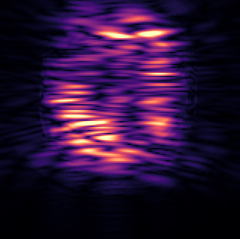} &
\includegraphics[width=0.115\linewidth]{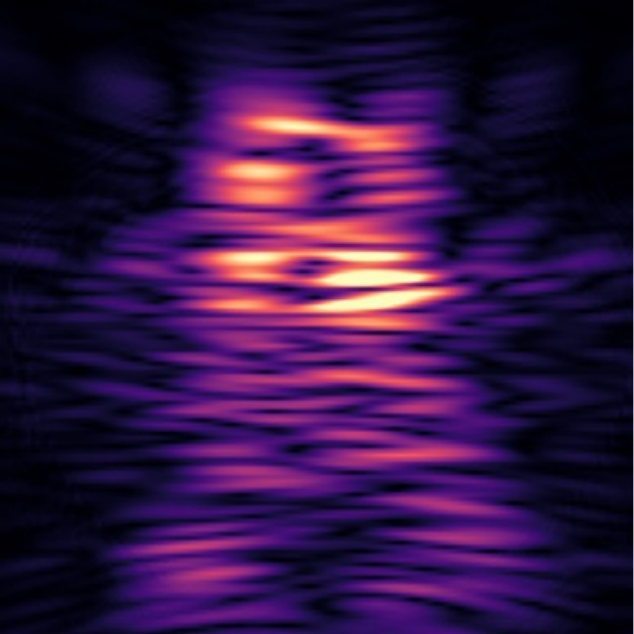} &
\includegraphics[width=0.115\linewidth]{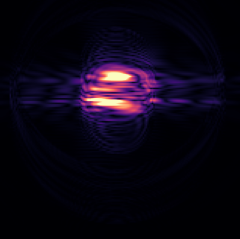} &
\includegraphics[width=0.115\linewidth]{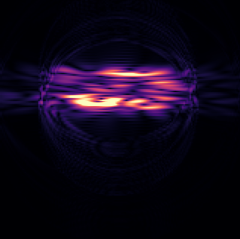} &
\includegraphics[width=0.115\linewidth]{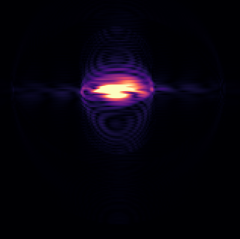} &
\includegraphics[width=0.115\linewidth]{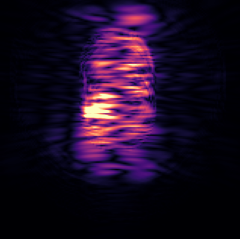} &
\includegraphics[width=0.115\linewidth]{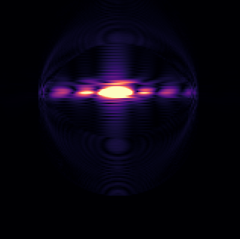} \\

\end{tabular}

\caption{Held-out synthetic reconstructions under extreme 1-scan sparsity.
Columns correspond to representative test objects excluded from the training split; rows show (top to bottom) ground-truth point clouds, SPIRIT-predicted geometry, sparse SAR observations, and predicted renderings. These examples visualize geometry recovery in the controlled synthetic setting; held-out physical-measurement results are reported separately in Appendix~\ref{sec:real_world}.}
\label{fig:qualitative_objects}
\end{figure*}

\section{Experimental Setup}
\label{sec:experiments}

\textbf{Dataset and Measurements.}
Our primary controlled evaluation uses synthetic sparse-aperture SAR measurements, for which both reference geometry and full-aperture coherent measurements are available.
After filtering incomplete instances, the benchmark contains 1,957 object--measurement instances, partitioned into 1,567 training, 195 validation, and 195 test instances; test objects are disjoint from the training objects.
We simulate a planar 77\,GHz FMCW radar operating over $77.5$--$80.5$\,GHz, with center frequency $79.0$\,GHz, bandwidth $3.0$\,GHz, and $512$ samples per chirp.
The synthetic aperture covers $200\,\mathrm{mm}\times100\,\mathrm{mm}$ and contains 13 scan lines as the full-aperture reference.
Sparse inputs retain 1, 3, or 6 scan lines and are represented as $\mathbf{Y}_{\mathrm{in}}=\mathbf{Y}_{\mathrm{full}}\odot\mathbf{M}$, where $\mathbf{M}$ is a binary scan-line mask and missing entries are zero-padded.
Training uses randomized scan-line masks, whereas validation and test results use fixed center-line masks for reproducible comparison. The synthetic data generator and SPIRIT renderer are distinct implementations: the generator operates on reference meshes with hard visibility and prescribed scattering components, whereas SPIRIT renders from its predicted structured surface with soft visibility and learned reflectivity/roughness. Both retain first-order geometry-based approximations, so the synthetic benchmark is best interpreted as a controlled attribution setting rather than complete evidence of real-world electromagnetic generalization.
Appendix~\ref{sec:real_world} evaluates a 60-object MITO subset with an object-disjoint 40/10/10 split; both captures of each object stay in one partition. Stage~I uses synthetic measurements only, and the disjoint real training split adapts only the Stage~II residual branch, so the full protocol is not zero-shot. Registered reference geometry is shared across methods without method-specific post-alignment, and the held-out physical test split uses geometry-, signal-, and image-domain metrics.

\textbf{Matched Baselines.}
We compare against backprojection, FISTA compressed sensing, an image-domain U-Net, a height-map prior, and implicit/volumetric neural-field controls using identical measurements, splits, masks, reconstruction volume, and metrics. FISTA uses 20 iterations ($\lambda=0.01$, step $0.5$), while the neural-field controls use a fixed 100-iteration per-scene budget; settings are fixed across test objects with no test-set tuning. These are matched controls rather than exhaustive reproductions of recent radar systems with different inputs or forward operators (Appendix~\ref{app:baseline}).

\textbf{Implementation Details.}
Differentiable backprojection converts sparse complex measurements into real, imaginary, and magnitude feature channels. An asymmetric ResNet--UNet predicts a $128\times128$ structured surface through the global--local parameterization in Section~4.2, and the renderer uses the single-bounce model in Section~4.3 with visibility sharpness $\kappa=10.0$. Full architecture details are provided in Appendix~\ref{app:experimental_details}.

\textbf{Training Protocol.}
Stage~I trains for 300 epochs with AdamW at $10^{-4}$; Stage~II jointly optimizes geometry, material parameters, and the refiner at $3\times10^{-6}$ with geometry regularization. Splits/masks use seed 42; Table~\ref{tab:quantitative_main} aggregates three independent SPIRIT runs, each on one RTX 8000 (Appendix~\ref{app:training}).

\textbf{Evaluation Metrics.}
We report CD (mm) for geometry, phase-history $\ell_1$ error for signal fidelity, and PSNR/SSIM on backprojected magnitude images. Geometry is the primary target; signal/image metrics are complementary. Definitions, forward projection for image-domain controls, and alignment procedures are given in Appendix~\ref{app:metrics}.

\begin{table*}[t]
\centering
\caption{Controlled synthetic comparison under 1-, 3-, and 6-scan sparsity. CD is reported in millimeters; PSNR/SSIM use backprojected magnitude images. SPIRIT is mean $\pm$ standard deviation over three seeds; controls use fixed schedules.}
\label{tab:quantitative_main}
\resizebox{\linewidth}{!}{
\begin{tabular}{lcccc|cccc|cccc}
\toprule
& \multicolumn{4}{c}{\textbf{1 Scan}} 
& \multicolumn{4}{c}{\textbf{3 Scans}} 
& \multicolumn{4}{c}{\textbf{6 Scans}} \\
\cmidrule(lr){2-5} \cmidrule(lr){6-9} \cmidrule(lr){10-13}

\textbf{Method}
& CD (mm) $\downarrow$ & Sig. Err. $\downarrow$ & PSNR $\uparrow$ & SSIM $\uparrow$
& CD (mm) $\downarrow$ & Sig. Err. $\downarrow$ & PSNR $\uparrow$ & SSIM $\uparrow$
& CD (mm) $\downarrow$ & Sig. Err. $\downarrow$ & PSNR $\uparrow$ & SSIM $\uparrow$ \\

\midrule
Backprojection
& 135.0 & \textbf{0.175} & 12.4 & 0.421
& 123.0 & \textbf{0.187} & 13.1 & 0.455
& 141.0 & 0.185 & 14.2 & 0.484 \\

Compressed Sensing 
& 82.0 & 0.208 & 13.5 & 0.421
& 64.0 & 0.208 & 12.5 & 0.427
& 55.0 & \textbf{0.162} & 13.6 & 0.422 \\

Implicit Neural Field
& 52.0 & 0.351 & 19.5 & 0.743
& 61.0 & 0.358 & 19.2 & 0.736
& 52.0 & 0.318 & 24.2 & 0.766 \\

Volumetric Neural Field
& 48.0 & 0.321 & 20.2 & 0.785
& 44.0 & 0.302 & 21.1 & 0.774
& 41.0 & 0.282 & 22.5 & 0.792 \\

Image-Domain UNet
& 75.0 & 0.684 & 12.4 & 0.681
& 63.0 & 0.512 & 19.1 & 0.705
& 52.0 & 0.556 & 19.3 & 0.721 \\

Height-Map Prior
& 81.0 & 0.324 & 20.5 & 0.521
& 57.0 & 0.256 & 21.7 & 0.621
& 32.0 & 0.232 & 23.3 & 0.675 \\

\midrule
\textbf{SPIRIT (Ours)}
& \textbf{23.1 $\pm$ 1.2} 
& 0.233 $\pm$ 0.015 
& \textbf{25.3 $\pm$ 0.3} 
& \textbf{0.829 $\pm$ 0.0063}
& \textbf{24.3 $\pm$ 1.5} 
& 0.242 $\pm$ 0.0032 
& \textbf{26.8 $\pm$ 0.4} 
& \textbf{0.842 $\pm$ 0.0039}
& \textbf{15.2 $\pm$ 1.9} 
& 0.235 $\pm$ 0.0042 
& \textbf{27.5 $\pm$ 0.3} 
& \textbf{0.854 $\pm$ 0.0041} \\

\bottomrule
\end{tabular}
}
\end{table*}

\subsection{Geometry Recovery under Sparse Aperture}
Table~\ref{tab:quantitative_main} evaluates the object-disjoint synthetic test set under 1-, 3-, and 6-scan inputs. SPIRIT obtains the lowest CD and highest PSNR/SSIM among the matched controls at all three sparsity levels. Its phase-history error is not uniformly lowest, showing that lower residual need not imply better geometry. Figure~\ref{fig:qualitative_objects} shows representative held-out 1-scan reconstructions; the matched representation controls contextualize the complete system without establishing universal optimality of fixed-topology surfaces.

\subsection{What Does Staged Optimization Contribute?}
Table~\ref{tab:ablation_core} isolates geometry initialization from phase-sensitive refinement under 1-scan sparsity. Full unfreezing without the refiner increases CD to $53.3$~mm despite higher PSNR, whereas the final \textit{Full Unfreeze + Refiner} configuration obtains CD $23.1$~mm, SSIM $0.8290$, PSNR $25.3$, and signal error $0.2050$. Thus, within the tested variants, structural regularization and phase-sensitive fitting play distinct roles, and measurement fit alone is not a reliable surrogate for scene correctness. Appendix~\ref{app:ablation} provides additional coverage, real-data, noise, and regularization analyses.

\paragraph{Real-world measurements.}
On object-disjoint MITO test objects after residual-only adaptation on a disjoint real split, SPIRIT has the lowest CD ($3.15$~mm) and highest PSNR/SSIM ($20.45$~dB/$0.805$) among matched controls; a volumetric neural field has a slightly lower signal residual ($0.353$ vs. $0.362$). This is a controlled near-field, non-zero-shot experiment (Appendix~\ref{sec:real_world}).

\begin{table}[t]
\centering
\caption{Core 1-scan ablation. Full unfreezing without the refiner causes geometric drift; the final joint configuration gives the best tested joint result.}
\label{tab:ablation_core}
\scriptsize
\setlength{\tabcolsep}{3.8pt}
\renewcommand{\arraystretch}{0.94}
\begin{tabular}{lcccc}
\toprule
\textbf{Configuration} & \textbf{CD (mm)} $\downarrow$ & \textbf{SSIM} $\uparrow$ & \textbf{PSNR} $\uparrow$ & \textbf{SigErr} $\downarrow$ \\
\midrule
Stage~I (Best) & 23.5 & 0.8273 & 17.1 & 0.2212 \\
Stage~I (Base) & 29.1 & 0.8110 & 19.2 & 0.2328 \\
Stage~II: Frozen Backbone + Refiner & 29.8 & 0.8150 & 21.3 & 0.2138 \\
Stage~II: Partial Geometry Freeze & 38.3 & 0.8070 & 20.3 & 0.2146 \\
Stage~II: Full Unfreeze (No Refiner) & 53.3 & 0.8160 & 22.1 & 0.2174 \\
Stage~II: Full Unfreeze + Refiner & \textbf{23.1} & \textbf{0.8290} & \textbf{25.3} & \textbf{0.2050} \\
\bottomrule
\end{tabular}
\renewcommand{\arraystretch}{1.0}
\end{table}

\section{Conclusion}
SPIRIT formulates severe sparse-aperture SAR as a non-injective inverse-imaging problem, combining a structured surface prior, differentiable RF rendering, and staged geometry--signal optimization. Controlled experiments show that lower measurement residual need not imply better geometry, motivating structural stabilization before phase-sensitive refinement; held-out MITO measurements provide complementary evidence after residual-only adaptation on a disjoint real split. The validated scope remains isolated, object-scale near-field sensing under approximate single-scattering physics, and the results support separating structural hypothesis selection from measurement refinement without claiming universal optimality of fixed topology.

\newpage
\bibliography{spirit_refs}
\bibliographystyle{spirit_refs}

\newpage
\appendix
\section{Appendix}
This appendix provides additional technical background, implementation details, and extended analyses that complement the main paper.
It is organized as follows:

\begin{itemize}
    \item \textbf{Appendix \ref{app:AA_radar_background}} reviews the radar and SAR fundamentals.
    \item \textbf{Appendix \ref{app:appendixB}} presents mathematical formulations, operator definitions, and optimization analysis of SPIRIT.
    \item \textbf{Appendix \ref{sec:preliminary}} describes preliminary experiments and motivating observations that inform the proposed formulation.
    \item \textbf{Appendix \ref{app:experimental_details}} provides comprehensive experimental details.
    \item \textbf{Appendix \ref{app:ablation}} reports extensive ablation studies and qualitative analyses.
    \item \textbf{Appendix \ref{app:discussion}} discusses limitations, future research directions, and broader implications.
    \item \textbf{Appendix \ref{app:ai-statement}} AI Use Statement.
\end{itemize}

\section{Extensive Radar and SAR Imaging Background}
\label{app:AA_radar_background}
\subsection{Electromagnetic Wave Propagation and Scattering Model}
\label{app:AA_em_model}

This appendix establishes the electromagnetic (EM) foundations underlying our radar forward model. 
We clarify the physical assumptions implicitly adopted by analytic RF rendering and delineate their validity regimes, which directly motivate the signal refinement strategy introduced in SPIRIT.

\subsubsection{From Maxwell’s Equations to the Helmholtz Equation}

Maxwell’s equations govern the propagation of electromagnetic waves. 
In a source-free, linear, isotropic, and non-dispersive medium, the time-domain equations are given by
\begin{equation}
\begin{aligned}
\nabla \times \mathbf{E}(\mathbf{x},t) &= -\mu \frac{\partial \mathbf{H}(\mathbf{x},t)}{\partial t}, \\
\nabla \times \mathbf{H}(\mathbf{x},t) &= \epsilon \frac{\partial \mathbf{E}(\mathbf{x},t)}{\partial t}, \\
\nabla \cdot \mathbf{E}(\mathbf{x},t) &= 0, \quad
\nabla \cdot \mathbf{H}(\mathbf{x},t) = 0,
\end{aligned}
\end{equation}
where $\mathbf{E}$ and $\mathbf{H}$ denote the electric and magnetic fields, and $\epsilon$ and $\mu$ are the permittivity and permeability of the medium.

Assuming time-harmonic fields of the form $\mathbf{E}(\mathbf{x},t) = \Re\{\mathbf{E}(\mathbf{x}) e^{j\omega t}\}$, Maxwell’s equations reduce to the frequency-domain vector Helmholtz equation,
\begin{equation}
\nabla^2 \mathbf{E}(\mathbf{x}) + k^2 \mathbf{E}(\mathbf{x}) = 0,
\end{equation}
where $k = \omega / c$ is the wavenumber and $c = 1/\sqrt{\mu\epsilon}$ is the speed of light. 
An analogous equation holds for the magnetic field. 
This formulation provides the starting point for most high-frequency radar imaging models.

\subsubsection{Single-Frequency Wave Propagation and Green’s Function}

For a monochromatic point source, the solution to the scalar Helmholtz equation can be expressed using the free-space Green’s function,
\begin{equation}
G(\mathbf{x},\mathbf{x}') = \frac{e^{-jk\|\mathbf{x}-\mathbf{x}'\|}}{4\pi\|\mathbf{x}-\mathbf{x}'\|},
\end{equation}
which describes spherical wave propagation from source location $\mathbf{x}'$ to observation point $\mathbf{x}$. 
In radar systems, both the transmitted and received fields can be modeled as superpositions of spherical waves, with the phase determined by the propagation distance. 
This distance-dependent phase term underlies coherent radar imaging and directly links geometry to the received signal.

\subsubsection{Scattering Model and Born Approximation}

To model the interaction between incident waves and a scene, we adopt a first-order single-scattering linearization,
in which the received field is modeled as a coherent superposition of independent surface-element contributions.
This Born-style assumption neglects higher-order multiple interactions, but preserves phase coherence and the geometric dependence of propagation.

Under a monostatic configuration, the complex-valued received signal at radar location $\mathbf{r}_m$ and wavenumber $k$
can be written in the same form as the main text as a discrete surface sum:
\begin{equation}
\hat{y}(\mathbf{r}_m,k)
\approx
\sum_{i,j}
\mathcal{V}(\mathbf{V}_{i,j},\mathbf{r}_m)\,
A(\mathbf{V}_{i,j})\,
\boldsymbol{\sigma}_{i,j}\,
\exp\!\big(-j\,2k\|\mathbf{r}_m-\mathbf{V}_{i,j}\|\big),
\end{equation}
where $\mathbf{V}_{i,j}$ and $\mathbf{N}_{i,j}$ denote surface vertices and normals,
$\mathcal{V}(\cdot)$ is a (differentiable) visibility term,
and $A(\cdot)$ absorbs system scaling and distance-dependent attenuation.
The complex scattering coefficient $\boldsymbol{\sigma}_{i,j}$ is parameterized by per-vertex reflectivity $\rho_{i,j}$
and roughness $\gamma_{i,j}$.

The Born approximation is valid when scattering is weak, and multiple reflections between scene elements can be neglected. 
This assumption typically holds for electrically small or moderately reflective objects in free space, and is commonly adopted in radar imaging literature for tractable modeling and inversion.

\paragraph{Limitations of the Single-Scattering Assumption}

While the single-scattering (single-bounce) surface-sum model enables efficient forward modeling and differentiable rendering, it does not fully capture the complexity of real-world RF signals. 
Its validity degrades in several common scenarios:
\begin{itemize}
    \item \textbf{Strong reflectors and specular surfaces}, where higher-order reflections dominate the measured signal.
    \item \textbf{Multipath propagation}, such as corner reflectors or urban environments, where waves undergo multiple bounces before returning to the sensor.
    \item \textbf{Material-dependent scattering and polarization effects}, which introduce angle- and frequency-dependent responses not captured by a scalar reflectivity model.
    \item \textbf{Hardware and calibration imperfections}, including antenna coupling and phase noise.
\end{itemize}

These effects introduce systematic discrepancies between analytic RF renderers and real signals. 
As a result, enforcing strict signal consistency with an idealized forward model can bias the reconstruction or amplify modeling errors, particularly under sparse-aperture sampling. 
This inherent model--reality mismatch motivates the signal refinement mechanism introduced in SPIRIT, which compensates for unmodeled scattering effects while preserving physical interpretability.

\subsubsection{FMCW Radar Signal Model}
\label{app:AA_fmcw}

Frequency Modulated Continuous Wave (FMCW) radar is widely used in modern robotic and automotive sensing systems due to its low power consumption, compact hardware, and simultaneous range and velocity estimation capability. 
This section formalizes the FMCW signal model and clarifies the physical meaning of the measured intermediate-frequency (IF) signal used throughout this work.

\subsubsection{Transmit Chirp Signal}

An FMCW radar transmits a continuous waveform whose instantaneous frequency increases linearly over time. 
The complex baseband representation of a single chirp can be written as
\begin{equation}
s_{\text{tx}}(t) = \exp\!\left(j 2\pi \left( f_c t + \tfrac{1}{2} K t^2 \right)\right), \quad t \in [0, T],
\end{equation}
where $f_c$ denotes the carrier frequency, $T$ is the chirp duration, and $K = B/T$ is the frequency sweep rate determined by the bandwidth $B$.

This linear frequency modulation encodes distance information as a frequency shift in the received signal, enabling efficient range estimation without pulsed transmission.

\subsubsection{Received Signal with Propagation Delay}

Consider a point scatterer located at range $R$ from the radar. 
The transmitted signal experiences a round-trip propagation delay $\tau = 2R/c$, yielding the received signal
\begin{equation}
s_{\text{rx}}(t) = \sigma\, s_{\text{tx}}(t - \tau)
= \sigma \exp\!\left(j 2\pi \left[ f_c (t-\tau) + \tfrac{1}{2} K (t-\tau)^2 \right]\right),
\end{equation}
where $\sigma$ denotes the complex reflectivity incorporating material response and propagation loss.

Direct digitization of $s_{\text{rx}}(t)$ would require sampling at the carrier frequency $f_c$, which is infeasible for mmWave radar systems due to analog-to-digital converter (ADC) bandwidth and power constraints.

\subsubsection{Dechirping and Intermediate-Frequency Signal}

To overcome this limitation, FMCW radar performs \emph{dechirping} by mixing the received signal with the conjugate of the transmitted signal,
\begin{equation}
s_{\text{IF}}(t) = s_{\text{rx}}(t)\, s_{\text{tx}}^*(t).
\end{equation}
Substituting the expressions above and neglecting higher-order terms in $\tau$, the resulting IF signal simplifies to
\begin{equation}
s_{\text{IF}}(t) \approx \sigma \exp\!\left(-j 2\pi K \tau t \right)\,
\exp\!\left(-j 2\pi f_c \tau \right).
\end{equation}

This operation shifts the signal from radio frequency to baseband while preserving phase coherence. 
The IF signal is a complex sinusoid whose frequency
\begin{equation}
f_b = K \tau = \frac{2 K R}{c}
\end{equation}
is directly proportional to the target range.

The use of the IF signal is essential: it reduces the required ADC sampling rate from the carrier frequency ($\sim$f$_c$) to the beat frequency ($\sim$MHz), while retaining both amplitude and phase information necessary for coherent imaging.
In our implementation, FMCW measurements provide a band of frequencies (equivalently, wavenumbers $k$); The forward model and losses are applied per-$k$ and aggregated across the band, matching the notation in the main text.

\subsubsection{Phase, Distance, and Geometric Sensitivity}
\label{app:AA_phase}

This section analyzes the relationship between phase, propagation distance, and geometric sensitivity in mmWave radar, providing the mathematical basis for the phase instability challenges encountered in sparse-aperture reconstruction.

\paragraph{Phase as a Function of Distance}

The round-trip propagation of a monochromatic wave induces a phase shift proportional to path length. 
For a scatterer at distance $R$, the accumulated phase is
\begin{equation}
\phi(R) = 2 k R = \frac{4\pi R}{\lambda},
\end{equation}
where $k = 2\pi/\lambda$ is the wavenumber and $\lambda$ denotes the carrier wavelength.

This expression highlights that the phase is not a nuisance term but a direct geometric encoding. 
Any change in distance $\Delta R$ induces a corresponding phase change
\begin{equation}
\Delta \phi = 2k\, \Delta R.
\end{equation}

\paragraph{Phase Sensitivity at Millimeter Wavelengths}

At millimeter-wave frequencies, wavelengths are on the order of a few millimeters. 
As a result, even sub-millimeter geometric perturbations produce significant phase variation. 
In particular,
\begin{equation}
\Delta R = \frac{\lambda}{4} \quad \Rightarrow \quad \Delta \phi = \pi,
\end{equation}
which corresponds to a complete sign inversion of the complex signal.

This extreme sensitivity implies that minor geometric errors can induce large oscillations in phase-based loss functions, yielding highly non-convex optimization landscapes with dense local minima. 
Such behavior is a fundamental property of coherent wave physics and cannot be eliminated solely through regularization or architectural design.

\paragraph{Phase is Structure, Not Noise}

Unlike incoherent imaging modalities, where phase is often discarded, phase in radar imaging encodes precise geometric structure. 
While noise may perturb phase signals, the underlying phase response remains deterministic and directly linked to propagation distance. 
Discarding or smoothing phase, therefore, removes essential structural information, undermining the ability to resolve fine geometry from coherent measurements.

\paragraph{Range Resolution and Bandwidth}

Range resolution in FMCW radar is determined by the signal bandwidth rather than carrier frequency. 
Specifically, the minimum resolvable distance difference is
\begin{equation}
\Delta R = \frac{c}{2B},
\end{equation}
where $B$ denotes the chirp bandwidth.

While increasing bandwidth improves range resolution, it does not mitigate phase-induced instability. 
This decoupling between range resolution and phase sensitivity underlies the ill-posed nature of sparse-aperture RF inverse imaging.

\subsection{Synthetic Aperture Radar Forward Model}
\label{app:AA_sar_forward}

Synthetic Aperture Radar (SAR) extends single-scan radar sensing by coherently combining multiple observations acquired from distinct spatial locations. 
This section formalizes the SAR forward model used throughout this work, explicitly defining the signal operator whose inversion under sparse sampling motivates our approach.

\subsubsection{Signal Notation}

We consider a monostatic SAR configuration in which a single radar platform serves as both transmitter and receiver. 
Let $\mathbf{r}_m \in \mathbb{R}^3$ denote the radar position at signal index $m$, corresponding to the $m$-th spatial location along the synthetic aperture trajectory.
The scene is modeled as a continuous spatial reflectivity function $\sigma(\mathbf{x}) \in \mathbb{C}$, where $\mathbf{x} \in \mathbb{R}^3$ denotes a spatial point.

Each SAR signal captures the coherent superposition of echoes scattered from all points in the scene under a single-scattering assumption.

\subsubsection{SAR Forward Operator}

For a given wavenumber $k = 2\pi/\lambda$, the received baseband signal at signal index $m$ can be expressed as
\begin{equation}
y_m(k)
\approx
\sum_{i,j}
\mathcal{V}(\mathbf{V}_{i,j},\mathbf{r}_m)\,
A(\mathbf{V}_{i,j})\,
\boldsymbol{\sigma}_{i,j}\,
\exp\!\big(-j\,2k\|\mathbf{r}_m-\mathbf{V}_{i,j}\|\big).
\label{eq:sar_forward}
\end{equation}

This expression accounts for the round-trip propagation phase accumulated between the radar position $\mathbf{r}_m$ and the scatterer at $\mathbf{x}$. 
Amplitude effects such as antenna gain and propagation loss are omitted here for clarity, as they do not alter the fundamental conditioning of the inverse problem.

Equation~(\ref{eq:sar_forward}) defines the single-scattering SAR forward operator $\mathcal{F}$ used in this work,
mapping the structured surface field $\mathcal{S}$ to complex measurements indexed by platform position and wavenumber.
\begin{equation}
\mathbf{y} = \mathcal{F}(\sigma).
\end{equation}

Each signal $y_m(k)$ is a global projection of the scene, and meaningful spatial discrimination emerges only through coherent aggregation across the synthetic aperture.

% \subsubsection{Coherent Stacking and Image Formation}

% Classical SAR image formation algorithms, such as backprojection, attempt to invert the forward model in~\eqref{eq:sar_forward} by coherently summing phase-compensated signals over the aperture. 
% When the set of radar positions $\{\mathbf{r}_m\}$ densely samples the aperture, this inversion is approximately stable. 
% However, when the aperture is sparse or incomplete, $\mathcal{F}$ becomes ill-conditioned with an ample null space, leading to fundamental ambiguities in reconstruction. 
% These properties are analyzed in detail in Appendix~A6.

\subsection{Resolution and Sparse Aperture Limitation}
\label{app:AA_limitation}

This section analyzes the fundamental resolution limits of SAR imaging and explains why sparse aperture sampling leads to non-injective inverse scattering.

\subsubsection{Rayleigh Resolution and Aperture Length}

In SAR, cross-range resolution is governed by the effective aperture length rather than pixel sampling density. 
Under standard assumptions, the Rayleigh cross-range resolution $\Delta x$ can be approximated as
\begin{equation}
\Delta x \approx \frac{\lambda R}{2L},
\end{equation}
where $\lambda$ is the wavelength, $R$ is the range to the target, and $L$ denotes the synthetic aperture length.

Increasing aperture length directly improves spatial resolution by increasing angular diversity and phase discrimination. 
This principle underlies the effectiveness of dense SAR trajectories for high-resolution imaging.

\subsubsection{Sparse Aperture Limitation}
Sparse-aperture sampling induces a range of characteristic artifacts arising from incomplete spectral coverage rather than noise or algorithmic deficiencies.

\paragraph{k-Space Coverage Interpretation}

From a spectral perspective, SAR signals correspond to samples of the scene’s spatial frequency content. 
Each radar position $\mathbf{r}_m$ contributes a distinct slice in spatial frequency (k-space), and dense aperture coverage fills this domain sufficiently to enable stable inversion.

Sparse-aperture sampling, however, results in incomplete and irregular k-space coverage. 
Large regions of the spatial frequency domain remain unobserved, resulting in information loss that cannot be recovered by post-processing.

\paragraph{Null Space and Non-Injectivity}

When aperture coverage is sparse, the SAR forward operator $\mathcal{F}$ admits a non-trivial null space. 
there exist distinct scene surfaces $\mathcal{S}_1 \neq \mathcal{S}_2$ such that
\begin{equation}
\mathcal{F}(\mathcal{S}_1) = \mathcal{F}(\mathcal{S}_2).
\end{equation}

This many-to-one mapping implies that the inverse problem is non-injective: no unique solution exists even in the absence of noise. 
As a result, sparse-scan SAR reconstruction is fundamentally ill-posed, and resolution cannot be recovered without introducing additional structural or physical constraints.

\paragraph{Grating Lobes and Ghost Targets}

Irregular or undersampled apertures produce grating lobes in the point spread function, causing energy from a single scatterer to replicate at multiple spatial locations. 
These ghost targets are coherent artifacts that persist even under noiseless signals.

\paragraph{Energy Leakage and Smearing}

Incomplete aperture coverage causes spectral leakage, spreading energy across the image domain and smearing fine structural details. 
This leakage masks weak scatterers and degrades contrast, particularly in high-dynamic-range scenes.

\paragraph{Spectral Leakage and Windowing}

Finite-aperture sampling induces spectral leakage in the spatial-frequency domain, leading to sidelobes in the reconstructed image. Windowing functions, such as Hamming or Kaiser windows, trade resolution for sidelobe suppression, thereby modifying the practical point spread function (PSF). In sparse-aperture scenarios, however, windowing alone cannot resolve ambiguities arising from missing spectral support.

\section{Mathematical and Algorithmic Details of SPIRIT}
\label{app:appendixB}

This appendix provides the mathematical formulations, operator definitions, and optimization analysis corresponding directly to the components introduced in Section~\ref{sec:method}.
% Appendix~A establishes the physical and signal-theoretic foundations of RF and SAR imaging.
% Appendix~B focuses on the differentiable structure, geometric parameterization, gradient behavior, and convergence rationale specific to the SPIRIT framework.

% ============================================================

\subsection{Differentiable Backprojection Encoder}
\label{app:AB_dbe}

SPIRIT employs a differentiable backprojection encoder that maps sparse, complex-valued SAR measurements into a spatially organized feature representation.
Backprojection can be understood as a per-location coherent accumulation process. For each spatial grid cell (or voxel) in the reconstruction volume, the encoder computes the expected propagation delay from every radar aperture position to that location, applies the corresponding phase compensation to the received complex signal, and coherently sums the contributions across apertures.
When a grid cell corresponds to a true scattering center, phase-aligned signals from multiple views constructively interfere, producing a strong response; otherwise, phase mismatch leads to destructive cancellation. In this way, the image is formed by aggregating energy independently at each spatial location.
Unlike classical backprojection used purely for image formation, our implementation preserves complex-valued phase information and is explicitly designed to be differentiable, allowing gradients to backpropagate through the coherent accumulation process and enabling end-to-end optimization under the SAR forward model.

\paragraph{Forward operator.}
Let $\mathbf{Y}_{\mathrm{obs}}=\{y_m(k)\}_{m\in\mathcal{M}}$ denote sparse FMCW SAR signals acquired at platform locations $\{\mathbf{r}_m\}_{m\in\mathcal{M}}$, indexed by wavenumber $k$.
For a candidate spatial location $\mathbf{x}\in\mathbb{R}^3$, the backprojection encoder is defined as
\begin{equation}
\mathcal{B}(\mathbf{x})
=
\sum_{m\in\mathcal{M}}
\int
y_m(k)\,
\exp\!\left(j\,2k\,\|\mathbf{r}_m-\mathbf{x}\|\right)
\, dk,
\label{eq:dbp_forward}
\end{equation}
which coherently aligns the phase histories associated with $\mathbf{x}$.
This operation can be interpreted as a matched filtering process in the spatial domain and corresponds to the adjoint of the SAR forward operator (Appendix~\ref{app:AA_sar_forward}).

In practice, the integral over $k$ is implemented via discrete summation, and signal values are obtained through differentiable interpolation when required.
The operator naturally supports arbitrary sparse apertures, as missing azimuth positions correspond to omitted summation terms.

\paragraph{Algorithmic evaluation.}
Algorithm~\ref{alg:dbp_forward} summarizes the forward evaluation of the differentiable backprojection encoder over a discrete spatial grid.

\begin{algorithm}[H]
\caption{Forward Differentiable Backprojection Encoder}
\label{alg:dbp_forward}
\begin{algorithmic}[1]
\REQUIRE Sparse signals $\{y_m(k)\}_{m\in\mathcal{M}}$, platform poses $\{\mathbf{r}_m\}$
\REQUIRE Spatial grid $\{\mathbf{x}_j\}_{j=1}^{N}$
\ENSURE Complex-valued feature map $\mathcal{B}(\mathbf{x})$
\FOR{each spatial location $\mathbf{x}_j$}
    \STATE $\mathcal{B}(\mathbf{x}_j) \leftarrow 0$
    \FOR{each signal index $m\in\mathcal{M}$}
        \FOR{each wavenumber sample $k$}
            \STATE $\phi_{mj}(k) \leftarrow \exp\!\left(j\,2k\,\|\mathbf{r}_m-\mathbf{x}_j\|\right)$
            \STATE $\mathcal{B}(\mathbf{x}_j) \mathrel{+}= y_m(k)\cdot \phi_{mj}(k)$
        \ENDFOR
    \ENDFOR
\ENDFOR
\STATE \textbf{return} $\mathcal{B}(\mathbf{x})$
\end{algorithmic}
\end{algorithm}

\paragraph{Differentiability and gradient flow.}
All operations in Eq.~(\ref{eq:dbp_forward}) are differentiable with respect to both the input signals and spatial coordinates.
Let $\mathcal{L}$ denote a scalar loss defined on downstream representations.
Gradients propagate through both the signal values and the geometric phase terms, yielding
\begin{equation}
\frac{\partial \mathcal{L}}{\partial y_m(k)}
=
\sum_{j}
\frac{\partial \mathcal{L}}{\partial \mathcal{B}(\mathbf{x}_j)}
\exp\!\left(j\,2k\,\|\mathbf{r}_m-\mathbf{x}_j\|\right),
\end{equation}
and
\begin{equation}
\frac{\partial \mathcal{L}}{\partial \mathbf{x}_j}
=
\sum_{m,k}
\frac{\partial \mathcal{L}}{\partial \mathcal{B}(\mathbf{x}_j)}
\, y_m(k)\,
j\,2k\,
\frac{\mathbf{x}_j-\mathbf{r}_m}{\|\mathbf{x}_j-\mathbf{r}_m\|}
\exp\!\left(j\,2k\,\|\mathbf{r}_m-\mathbf{x}_j\|\right).
\end{equation}
These expressions ensure that gradients explicitly respect wave-propagation geometry and phase coherence.

\paragraph{Complex feature construction.}
The backprojection output $\mathcal{B}(\mathbf{x})$ is complex-valued.
To interface with standard neural architectures while preserving coherent information, we construct a multi-channel real feature representation.
\[
\mathbf{Z}(\mathbf{x})
=
\big[
\Re\{\mathcal{B}(\mathbf{x})\},
\Im\{\mathcal{B}(\mathbf{x})\},
|\mathcal{B}(\mathbf{x})|
\big].
\]
This representation retains both phase-sensitive and magnitude-based cues, which are critical under sparse and non-injective collection regimes (Appendix~\ref{app:AA_limitation}).

\paragraph{Discussion.}
The differentiable backprojection encoder serves as a physics-aligned signal transform that bridges sparse SAR observations and spatially structured representations.
By exposing coherent phase alignment to gradient-based optimization, SPIRIT can reason directly in the space of physically realizable SAR images rather than relying on pixel-domain heuristics.

% ============================================================
\subsection{Structured Grid Scene Parameterization}
\label{app:AB_grid}

SPIRIT represents geometry as a structured $H\times W$ surface grid deformed into 3D space.

\paragraph{Bounding-box parameterization.}
Vertices are generated via
\begin{equation}
\mathbf{V}_{i,j}
=
\mathbf{v}_{\min}
+
\boldsymbol{\delta}_{i,j}
\odot
(\mathbf{v}_{\max}-\mathbf{v}_{\min}),
\qquad
\boldsymbol{\delta}_{i,j}\in[0,1]^3.
\end{equation}
This parameterization constrains all vertices to a compact spatial domain.

\paragraph{Stability under sparse inverse scattering.}
Under sparse aperture coverage, the SAR forward operator admits a large null space (Appendix~\ref{app:AA_limitation}).
Unbounded point-based representations can drift arbitrarily while still matching subsets of signals.
Bounding-box constraints eliminate scale-free degeneracies and stabilize optimization by restricting the hypothesis space to physically plausible geometry.

\paragraph{Why a structured grid.}
Unlike unstructured point clouds, the grid induces explicit topology and consistent neighborhood relations.
This provides consistent local neighborhoods for normal computation, surface orientation, and gradient propagation during wave-consistent rendering under sparse signals.

% ============================================================
\subsection{Normal Computation and Gradient Flow}
\label{app:AB_normals}

Surface normals are derived deterministically from the grid geometry.

\paragraph{Finite-difference formulation.}
For interior vertices,
\begin{align}
\mathbf{t}^x_{i,j} &= \mathbf{V}_{i+1,j}-\mathbf{V}_{i-1,j},\\
\mathbf{t}^y_{i,j} &= \mathbf{V}_{i,j+1}-\mathbf{V}_{i,j-1},
\end{align}
and the normal is
\begin{equation}
\mathbf{N}_{i,j}
=
\frac{\mathbf{t}^x_{i,j}\times \mathbf{t}^y_{i,j}}
{\|\mathbf{t}^x_{i,j}\times \mathbf{t}^y_{i,j}\|_2+\epsilon}.
\end{equation}

\paragraph{Differentiability.}
This construction is fully differentiable.
Gradients from the renderer propagate through normals to vertex positions, encouraging locally smooth and physically consistent surface orientation.
Learning normals independently was empirically unstable and produced non-integrable surfaces.

% ============================================================
\subsection{Differentiable RF Renderer}
\label{app:AB_renderer}

The renderer implements a discrete, differentiable approximation of the SAR forward operator (Appendix~\ref{app:AA_sar_forward}).

\paragraph{Signal synthesis.}
For platform position $\mathbf{r}_m$ and wavenumber $k$, the synthesized signal is
\begin{equation}
\hat{y}_m(k)
=
\sum_{i,j}
A(\mathbf{V}_{i,j},\mathbf{r}_m,k)\,
\sigma_{i,j}(k)\,
\eta_{m,i,j}
\,
\exp\!\big(-j\,2k\,\|\mathbf{r}_m-\mathbf{V}_{i,j}\|\big),
\end{equation}
where $A(\cdot)$ models distance-dependent attenuation and known system scaling, while $\sigma_{i,j}(k)$ denotes the complex scattering coefficient parameterized by the predicted material descriptors. In our implementation, the attenuation term includes a $1/R$ decay factor, consistent with the experimental setup.

\paragraph{Soft visibility.}
Visibility is approximated as
\begin{equation}
\eta_{m,i,j}
=
\sigma\!\big(\kappa\,\langle \mathbf{N}_{i,j}, \mathbf{d}_{m,i,j}\rangle\big),
\end{equation}
where $\mathbf{d}_{m,i,j}$ is the view direction.
This soft formulation avoids zero-gradient discontinuities and provides continuous orientation cues for optimization.

\paragraph{Geometric gradients.}
The full renderer gradient contains contributions from distance attenuation, soft visibility, surface normals, and phase. The phase-dependent component for a single surface contribution $\hat{y}_{m,i,j}(k)$ is proportional to
\begin{equation}
\left.\frac{\partial \hat{y}_{m,i,j}(k)}{\partial \mathbf{V}_{i,j}}\right|_{\mathrm{phase}}
=
-j2k\,\hat{y}_{m,i,j}(k)
\frac{\mathbf{V}_{i,j}-\mathbf{r}_m}{\|\mathbf{V}_{i,j}-\mathbf{r}_m\|},
\end{equation}
with additional terms from the geometry dependence of attenuation and visibility. Automatic differentiation uses the complete renderer; the displayed phase term isolates the rapidly oscillatory component that drives the non-convexity discussed in Appendix~\ref{app:AA_limitation}.

% ============================================================
\subsection{Signal Refinement as Constrained Residual Learning}
\label{app:AB_refiner}

Analytic RF rendering cannot capture all real-world effects (Appendix~\ref{app:AA_limitation}).
SPIRIT introduces a signal refinement module to compensate for this model--reality mismatch.

\paragraph{Residual formulation.}
Given analytically rendered signals $\hat{\mathbf{Y}}$, the refined prediction is
\begin{equation}
\tilde{\mathbf{Y}}
=
\hat{\mathbf{Y}}
+
\beta\,\mathcal{G}_{\psi}(\hat{\mathbf{Y}}),
\end{equation}
where $\mathcal{G}_{\psi}$ is a lightweight residual network and $\beta$ is initialized small.

\paragraph{Identity regularization and schedule.}
The refiner is regularized toward a near-identity correction and is activated only after coarse geometry convergence. The residual branch receives the analytically rendered signal rather than scene parameters and does not provide a direct residual-gradient path to the geometry encoder. This separation limits the extent to which residual learning can mask geometric error, but it is not a hard physical guarantee: the refiner may still absorb signal structure omitted by the approximate renderer.

% ============================================================
\subsection{Optimization Landscape and Two-Stage Training}
\label{app:AB_optimization}

\paragraph{Phase-induced non-convexity.}
The RF phase $\phi = 2kR$ implies that a displacement $\Delta R = \lambda/4$ induces a $\pi$ phase shift (Appendix~\ref{app:AA_phase}).
This produces a highly oscillatory loss surface with many local minima and cycle-skipping behavior.

\paragraph{Continuation interpretation.}
Let
\[
\mathcal{L}(\mathbf{V})
=
\mathcal{L}_{\mathrm{mag}}(\mathbf{V})
+
\lambda\,\mathcal{L}_{\mathrm{phase}}(\mathbf{V}).
\]
Stage~I sets the weight of phase-sensitive terms to zero, with a smoothed objective dominated by magnitude- and geometry-based losses.
Stage~II increases $\lambda$ after the geometry estimate has entered a stable local basin, reintroducing phase-sensitive consistency only after coarse structure has been established. This continuation-style strategy reduces the tendency of early phase gradients to drive large geometric errors under sparse apertures; it does not imply convergence to a unique or global optimum.

\section{Preliminary}
\label{sec:preliminary}

\begin{figure}[ht]
   \centering
   \begin{subfigure}[b]{0.35\textwidth}
       \centering\includegraphics[width=\textwidth]{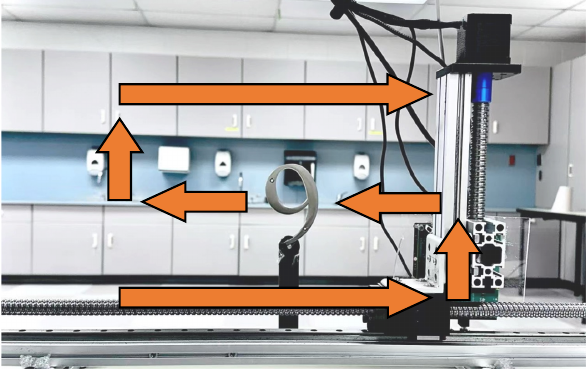}
       \caption{Data Collection Setup}
   \end{subfigure}
   \begin{subfigure}[b]{0.35\textwidth}
       \centering\includegraphics[width=\textwidth]{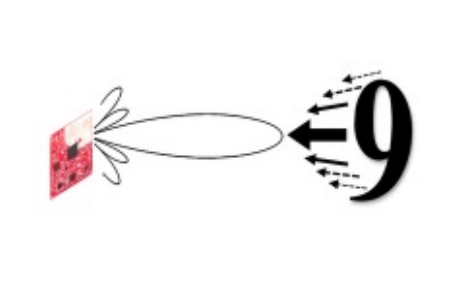} 
       \caption{Propagation Illustration}
   \end{subfigure}
   \caption{Real-world SAR data collection and signal propagation illustration.
(a) Physical scanning setup and Z-shaped raster trajectory used for planar SAR collection, consistent with the simulator configuration.
(b) Conceptual illustration of wave propagation and angle-dependent scattering, highlighting that while the beam covers the entire region of interest, different aperture locations emphasize different geometric components of the target.}

    \label{fig:pre_explain}
\end{figure}

\begin{figure*}[ht]
   \centering
   \begin{subfigure}[b]{0.2\textwidth}
       \centering\includegraphics[width=\textwidth]{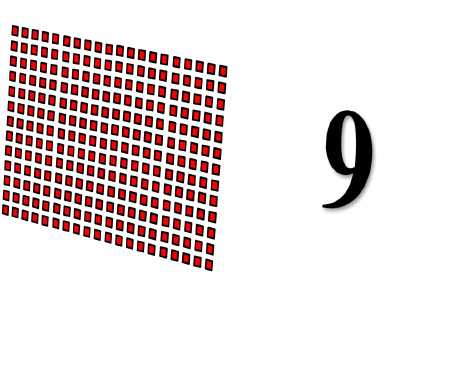}
       \caption{Full Radar}
   \end{subfigure}
   \hfill
   \begin{subfigure}[b]{0.2\textwidth}
       \centering\includegraphics[width=\textwidth]{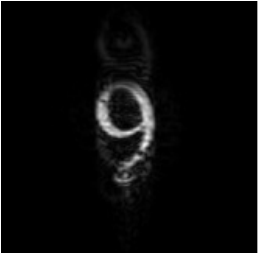} 
       \caption{Full SAR}
       \label{fig:full_sar}
   \end{subfigure}
   \hfill
      \begin{subfigure}[b]{0.2\textwidth}
       \centering\includegraphics[width=\textwidth]{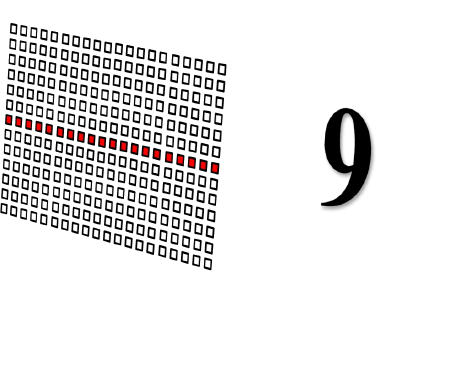}
       \caption{Sparse Radar}
   \end{subfigure}
   \hfill
   \begin{subfigure}[b]{0.2\textwidth}
       \centering\includegraphics[width=\textwidth]{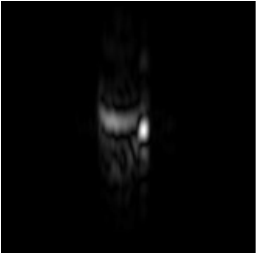} 
       \caption{Sparse SAR}
       \label{fig:sparse_sar}
   \end{subfigure}
   
      \begin{subfigure}[b]{0.2\textwidth}
       \centering\includegraphics[width=\textwidth]{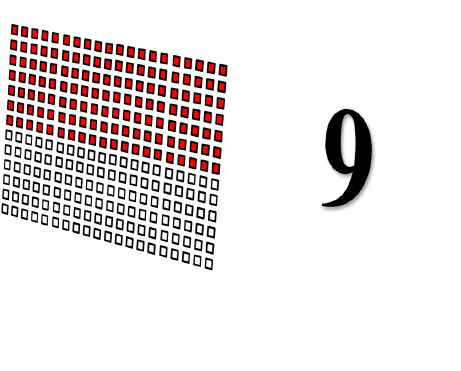}
       \caption{Upper Radar}
   \end{subfigure}
   \hfill
   \begin{subfigure}[b]{0.2\textwidth}
       \centering\includegraphics[width=\textwidth]{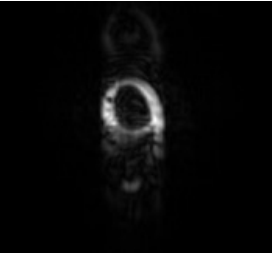} 
       \caption{Upper SAR}
       \label{fig:up_sar}
   \end{subfigure}
   \hfill
      \begin{subfigure}[b]{0.2\textwidth}
       \centering\includegraphics[width=\textwidth]{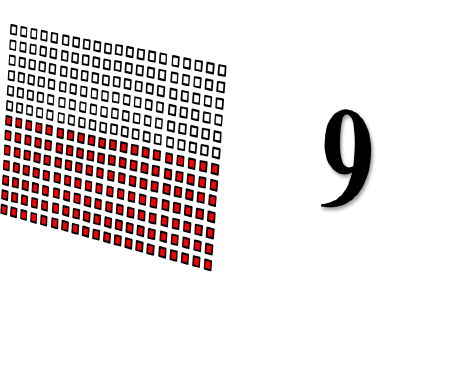}
       \caption{Lower Radar}
       \label{fig:low_sar}
   \end{subfigure}
   \hfill
   \begin{subfigure}[b]{0.2\textwidth}
       \centering\includegraphics[width=\textwidth]{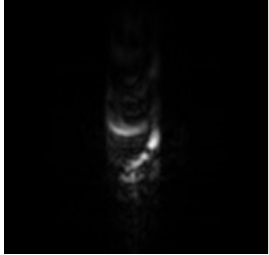} 
       \caption{Lower SAR}
   \end{subfigure}

\caption{Real-world SAR observations of a metallic digit 9 under different aperture configurations.
Black boxes mark radar aperture locations, and the red marker denotes the active position.
From left to right: full-aperture, single-pass sparse aperture, upper-aperture subset, and lower-aperture subset.
Full-aperture integration produces the most complete formed image, whereas restricted aperture subsets emphasize different portions of the target. The comparison illustrates viewpoint-dependent scattering and the loss of angular/spatial-frequency support under sparse acquisition.
}

    \label{fig:pre_result}
\end{figure*}

We begin with a real-world SAR collection to motivate the sparse-aperture setting and illustrate how aperture position changes the formed image under a fixed field of view. This preliminary study is qualitative: it demonstrates viewpoint-dependent scattering and incomplete angular support, rather than claiming that sparse sampling preserves all information available from the dense aperture. Figures~\ref{fig:pre_explain}--\ref{fig:pre_result} use conventional formed SAR observations for problem motivation; they are not final SPIRIT reconstructions and are not part of the quantitative MITO evaluation in Appendix~\ref{sec:real_world}.
% Establishing this observation on real data is critical before introducing any reconstruction model.

Data are collected using a Commercial Off-the-Shelf (COTS) mmWave radar, the TI IWR 1642, with a mechanical setup comprising the SAR setup shown in Figure \ref{fig:pre_explain}. 
% whose geometry and collection parameters exactly match those of our simulator.
The radar performs a planar scan over a rectangular region of size $200\,\mathrm{mm} \times 100\,\mathrm{mm}$ following a Z-shaped raster trajectory, repeated 13 times.
A metallic digit 9 is placed within the scanning region as the target.
The target remains fully illuminated throughout the collection, as the radar beamwidth covers the entire region of interest at all aperture locations.
This setup keeps the target within the nominal field of view across the scan, allowing us to separate field-of-view coverage from the loss of angular and spatial-frequency support caused by using fewer aperture positions.

The aggregated reconstruction obtained by coherently combining all 13 scans is shown in Fig.~\ref{fig:full_sar}.
With dense aperture coverage, the characteristic geometry of the digit 9 becomes clearly visible.
% confirming that the object structure is physically encoded in the measured radar signals.
% This result highlights that high-resolution geometric information emerges when angle-dependent scattering contributions are coherently integrated across the aperture.
However, a reconstruction from a single scan, shown in Fig.~\ref{fig:sparse_sar}, produces incomplete and spatially ambiguous structures.
The single-scan reconstruction is incomplete because the restricted aperture provides substantially less angular and spatial-frequency support than the full collection. The backscattered energy is also highly viewpoint dependent. 
Only surfaces that are specularly aligned with the viewing direction produce strong coherent returns; off-angle regions are dominated by diffuse scattering with substantially weaker backscattered energy.
These weak responses cause structural details to be blurred or missed.

To explicitly demonstrate this effect, we design a controlled experiment that isolates the contribution of aperture position.
We separately reconstruct SAR images from the upper and lower six scans.
As shown in Fig.~\ref{fig:up_sar}, radar positions are illustrated as black rectangles, and the subset highlighted in red corresponds to the aperture locations used to generate the image.
The upper-aperture subset produces an image that clearly resolves the lower portion of the digit 9.
The image generated from the lower-aperture subset in Fig.~\ref{fig:low_sar} reveals the upper portion of the digit.
Although the reflected intensity is weaker in this case, the global shape remains partially recognizable. Together, these observations show that different aperture subsets contribute complementary measurements of the same object, while no individual subset provides the full support available from the dense aperture.
\section{Experimental Details}
\label{app:experimental_details}

This appendix records the experimental settings used to reproduce the reported comparisons, including data splits, simulation assumptions, preprocessing, baseline definitions, optimization schedules, compute, and metrics.

\subsection{Datasets}

\paragraph{Synthetic corpus.}
After filtering incomplete object--measurement instances, the synthetic corpus contains 1,957 valid instances: 1,567 for training, 195 for validation, and 195 for testing. The partitions are object-disjoint at test time. Synthetic scenes cover heterogeneous object geometry, orientation, and surface structure.

\paragraph{MITO real measurements.}
We evaluate a 60-object subset of the MITO dataset~\cite{dodds2025mito}. The real-data split is object-disjoint: 40 objects are used for Stage~II residual adaptation, 10 for validation, and 10 for testing. Each object has two captures at horizontal orientations separated by $90^\circ$, and both captures of the same object remain in the same partition, yielding 80/20/20 train/validation/test captures. Stage~I geometry learning uses no real MITO measurements; the real training split is used only to adapt the residual-refinement branch. Thus, the real test captures are held out from real-data adaptation, although the experiment is not presented as object-identity zero-shot with respect to the broader synthetic mesh corpus. The released complex-valued 77\,GHz measurements and recorded radar poses are processed using the dataset trajectory; the physical measurements therefore include hardware response, positioning/calibration residuals, material-dependent scattering, and multipath effects not reproduced exactly by the synthetic generator or analytic renderer. Reference geometry is registered to the radar coordinate frame using the dataset calibration/object pose. The same aligned reference is used for all methods, with no method-specific post-hoc alignment before Chamfer Distance evaluation.

\subsection{Radar Signal Simulation}

\paragraph{Waveform Parameters.}
We simulate a 77\, GHz FMCW radar, closely modeled after Texas Instruments mmWave sensors.
The operating frequency range is $77.5$--$80.5$\,GHz, with center frequency $f_c = 79.0$\,GHz and bandwidth $B = 3.0$\,GHz.
The chirp slope is $S = 5.86 \times 10^{12}$\,Hz/s, yielding a chirp duration
$T_c = B/S \approx 512\,\mu$s.
Each chirp contains $N_s = 512$ ADC samples, corresponding to a sampling rate of approximately $1$\, MHz.
The theoretical range resolution is $\delta_r = c/(2B) \approx 5$\,cm.

\paragraph{Signal Processing.}
Raw ADC signals are converted to the frequency domain by applying a one-dimensional Fast Fourier Transform along the fast-time dimension.
A Hanning window is applied before the FFT to suppress sidelobes.
All subsequent reconstruction, rendering, and learning-based processing operate on the resulting complex-valued range-frequency representation.

\paragraph{Forward Scattering Model.}
The synthetic measurement generator is distinct from the differentiable renderer optimized by SPIRIT. The generator operates on reference meshes with hard visibility/ray casting and prescribed scattering components, whereas SPIRIT renders from its predicted structured surface using soft visibility and learned reflectivity/roughness. Both are geometry-based first-order approximations and neither models all higher-order multi-bounce effects; the synthetic benchmark is therefore partially model-matched at the level of physical assumptions, but not an inversion of an identical renderer.
For radar position $\mathbf{r}_m$ and wavenumber $k$, the simulated return is
\begin{equation}
y_m(k) =
\sum_{i \in \mathcal{V}_{\mathrm{vis}}}
A(\|\mathbf{r}_m-\mathbf{v}_i\|)\,\rho_i\,
\exp\!\big(-j\,2k\,\|\mathbf{r}_m-\mathbf{v}_i\|\big),
\end{equation}
where $\mathbf{v}_i$ denotes a visible mesh vertex, $\rho_i$ is its reflectivity, and $A(\cdot)$ captures system scaling and distance-dependent attenuation.

% \paragraph{Scatterers and Noise Model.}
% Scatterers correspond to the vertices of high-resolution object meshes.
% The core simulation isolates geometric reconstruction errors from signal-to-noise limitations.
% We adopt this setting to isolate the effect of aperture sparsity; robustness to noise is discussed as a limitation and left for future work.
% We emphasize that our core claim concerns the ill-posedness induced by aperture sparsity (non-injectivity) rather than measurement noise.

\paragraph{Visibility Culling.}
Visible scatterers $\mathcal{V}_{\mathrm{vis}}$ are determined via a two-stage culling process.
First, backface culling removes vertices whose surface normals $\mathbf{n}_i$ satisfy
$\mathbf{n}_i \cdot (\mathbf{p} - \mathbf{v}_i) \leq 0$.
Second, exact ray-casting is performed to discard vertices occluded by other parts of the mesh.
This ensures that only physically visible surface elements contribute to the simulated return.

\subsection{Synthetic Aperture Geometry}

\paragraph{Aperture Configuration.}
Radar measurements are collected over a planar rectangular synthetic aperture with dimensions approximately $200$\, mm $\times$ $100$\, mm.
The aperture is sampled on a dense Cartesian grid, satisfying the spatial Nyquist criterion and preventing grating lobes.

% \paragraph{Scanning Protocol.}
% The radar follows a deterministic raster-scanning pattern across the aperture, traversing rows sequentially in alternating scan directions.
% The radar plane is fixed at $z = 0$ throughout collection.
% All objects are located at positive depth values $z \in [0.15, 0.30]$\,m, where $z$ denotes distance from the radar plane.

\subsection{Sparse Aperture Sampling}

\paragraph{Measurement Representation.}
Full-aperture measurements are represented as a complex-valued tensor
$\mathbf{Y}_{\mathrm{full}} \in \mathbb{C}^{N_y \times N_x \times N_s}$,
where $N_x$ and $N_y$ denote aperture positions and $N_s$ denotes range-frequency bins.
% Sparse observations are defined by a binary mask
% $\mathbf{M} \in \{0,1\}^{N_y \times N_x}$.
In our datasets, scanline sparsity corresponds to selecting a subset of $N_y$ rows (vertical scan positions) while retaining all $N_x$ columns within each selected row.

\paragraph{Sampling Strategy.}
We distinguish between training-time and evaluation-time sparsity patterns.
During training, we apply random scanline masks as data augmentation,
randomly selecting subsets of scan lines to improve robustness to sampling variations.
For validation and all reported test results, we use fixed and deterministic center scanline masks,
corresponding to the central 1, 3, or 6 scan positions described in Section 5 of the main paper.
All baseline methods are evaluated under the identical sparsity masks
to ensure a fair and controlled comparison.
All quantitative results are computed using these fixed-center scanline masks.

\paragraph{Training and Evaluation Protocol.}
During training, a new sparse mask is sampled independently for each mini-batch to prevent overfitting to any fixed sampling pattern.
During validation and testing, sparse masks are fixed to ensure reproducible and fair comparisons across methods.
The Hadamard product gives the network input
\begin{equation}
    \mathbf{Y}_{\mathrm{in}} = \mathbf{Y}_{\mathrm{full}} \odot \mathbf{M},
\end{equation}
with missing measurements zero-padded.
The masked signal $\mathbf{Y}_{\mathrm{in}}$ is used as input to the differentiable backprojection operator,
whose output feature map serves as the direct input to the reconstruction backbone (Section~4.2).
To preserve expected signal energy across sparsity levels, non-zero measurements are normalized by the inverse sparsity ratio.
This normalization is applied consistently to all methods, including baselines.

\subsection{Synthetic Scene Generation}

\paragraph{Object Assets.}
Object meshes are sourced from the MuJoCo physics engine and the MITO object dataset.
All meshes are preprocessed to ensure watertightness, metric scaling, and centered geometry.

\paragraph{Diversity and Randomization.}
Scenes include both simple geometric primitives and complex household objects.
Each object undergoes random 3D rotation (roll, pitch, yaw uniformly sampled in $[0, 2\pi]$) and slight translational jitter of up to $\pm 2$\, cm.
Test scenes contain object instances that were not seen during training.

\subsection{Model Architecture}

\paragraph{Geometry Network.}
The geometry reconstruction network follows an asymmetric ResNet--UNet architecture.
The encoder comprises four stages with channel sizes $\{64, 128, 256, 512\}$, followed by a self-attention bottleneck.
The decoder mirrors the encoder with skip connections.
The output is a $128 \times 128$ structured surface grid of 3D vertex coordinates.

\paragraph{Parameterization.}
Geometry is parameterized using a global--local formulation.
A global MLP predicts an axis-aligned bounding box, while convolutional heads predict normalized local offsets within this box.

\paragraph{Physics-Based Renderer.}
The VertexRenderer module is fully differentiable with respect to vertex positions.
Visibility is modeled using a soft normal-based gating function
$\sigma(\kappa\,\mathbf{n}\cdot\mathbf{v})$,
which provides stable gradients for surface orientation during training.

\subsection{Data Representation and Preprocessing}
\paragraph{Input Signal Representation.}
The overall input to SPIRIT is the sparsely sampled complex-valued SAR signal.
This signal is first transformed through a differentiable backprojection operator,
which aggregates the sparse measurements into a spatial feature representation.
The resulting BP feature map serves as the direct input to the UNet-style reconstruction backbone.
The sparse complex signal is not fed directly into the UNet; instead, it is implicitly encoded by the BP operator.
Complex signals are represented using real and imaginary components, stored as complex64 tensors in PyTorch.

\paragraph{Log-Magnitude Normalization.}
To stabilize training and reduce dynamic range, the input signal magnitude is normalized to the log-magnitude.
Specifically, given a complex signal $S$:
\begin{equation}
m = 20\log_{10}(|S|+\epsilon), \qquad
m_{\mathrm{clip}}=\mathrm{clip}(m,-100~\mathrm{dB},0~\mathrm{dB}),
\end{equation}
\begin{equation}
S_{\mathrm{norm}}=\mathrm{MinMax}(m_{\mathrm{clip}})\cdot e^{j\angle S}.
\end{equation}
Unobserved measurements are set to zero, while the mask $M$ restricts network attention and backprojection to valid samples.

\paragraph{Reconstruction Volume.}
The reconstruction volume is defined in the physical coordinate frame specified by the simulator configuration.
Typical bounds are $[-0.15, 0.15]$~m per axis for validation crops, and up to $[-0.3, 0.3]$~m for full-field simulation.
Voxel resolution is $128^3$ or $256^3$, depending on the experiment.
These volumetric bounds and resolutions are used for simulation and volumetric baselines; SPIRIT itself predicts a structured surface grid.

\paragraph{Sparse Sampling Strategy.}
Sparsity is structured by scan lines rather than random point dropping.
The dataset selects a subset of vertical azimuth scan lines and masks all remaining positions.
This setting reflects realistic sparse-aperture collection and preserves angular coherence.

\subsection{Baseline Implementations}
\label{app:baseline}

All baselines operate on the exact sparse measurements and are evaluated under identical conditions. They are designed as matched controls under our acquisition model; the implicit and volumetric variants are not presented as faithful reimplementations of Radar Fields, GeRaF, SpINRv2, RadarSplat, or other systems whose inputs, forward operators, and scene regimes differ from ours.

\paragraph{Backprojection.}
The BP baseline implements a classical time-domain SAR backprojection using frequency-domain phase correction.
Given a radar location $\mathbf{r}$ and spatial location $\mathbf{x}$, the contribution is:
\begin{equation}
I(\mathbf{x})=\left|\sum_{m\in\mathcal{M}}\sum_{k} y_m(k)\exp\!\big(j\,2k\|\mathbf{x}-\mathbf{r}_m\|\big)\right|,
\end{equation}
where $y_m(k)$ denotes the measured frequency-domain signal.
The output is the magnitude of the complex image.

\paragraph{Compressed Sensing (CS).}
The CS baseline solves a sparse inverse problem:
\begin{equation}
\min_{\mathbf{x}} \| \mathbf{y} - \mathbf{A}\mathbf{x} \|_2^2 + \lambda \|\mathbf{x}\|_1,
\end{equation}
where $\mathbf{x}$ is a volumetric reflectivity grid and $\mathbf{A}$ denotes the forward operator.
Optimization is performed using FISTA for the fixed 20-iteration schedule reported in the main paper, with $\lambda=0.01$ and step size $0.5$. The same values are used for every test object and sparsity level, with no per-test-instance tuning. A common fixed threshold is applied to extract surfaces for geometric evaluation.

\paragraph{Image-Domain UNet.}
The UNet baseline takes a coarse BP image as input.
Three channels are provided: real, imaginary, and magnitude components.
The architecture matches the SPIRIT backbone but operates purely in the image domain.

\paragraph{Height Baseline.}
The height baseline uses the same backbone as SPIRIT but assumes a fixed planar surface at the mid-depth of the volume.
Only reflectivity is predicted on this surface; no 3D deformation is allowed.
This baseline isolates the benefit of explicit surface modeling.

\paragraph{Implicit and Volumetric Baselines.}
Implicit and volumetric neural-field controls optimize a representation independently for each test sample. They do not generalize across the dataset. Each sample uses the same fixed 100-iteration budget and minimizes the L2 discrepancy between predicted and observed signals. Appendix tables additionally retain a direct volumetric variant from the same matched benchmark pipeline. These controls are included to probe representation behavior under a common acquisition/evaluation protocol, not as faithful reproductions of named published neural-field systems.

\subsection{Training Protocol}
\label{app:training}
\paragraph{Stage~1 Training.}
Stage~I focuses on stabilizing geometric reconstruction. During this stage, supervision is limited to geometry-related regularization and magnitude-based or image-domain losses derived from backprojected measurements. No complex-valued phase supervision or signal-level matching is enforced in Stage~I. We use AdamW for 300 epochs with learning rate $10^{-4}$, weight decay $10^{-4}$, cosine annealing to $10^{-6}$, gradient clipping at $0.5$, batch size $8$, and gradient accumulation over $8$ steps.

\paragraph{Stage~2 Training.}
Stage~2 introduces complex-valued signal supervision and activates the residual refinement module. In the final SPIRIT configuration, the geometry backbone remains trainable but uses the reduced Stage~II learning rate together with geometry-preserving regularization; material parameters and the refiner are optimized jointly. Frozen-backbone, partial-freeze, and no-refiner variants are used only as controlled ablations in Table~\ref{tab:ablation_core}. The optimizer and learning-rate scheduler are reinitialized at the beginning of Stage~2.

% Training uses AdamW with weight decay $10^{-4}$ and a cosine annealing schedule.
% Batch size is 2 per GPU, using 8 GPUs.
% Stage~2 is trained for 100 epochs with a base learning rate of $10^{-5}$.

\subsection{Computational Resources}
We report the compute resources used for training and evaluation to improve reproducibility. All experiments were run on a machine with Intel(R) Xeon(R) Platinum 8268 CPUs and NVIDIA Quadro RTX 8000 GPUs with 48\,GB VRAM. Unless otherwise noted, each experiment uses a single RTX 8000 GPU, although the full server includes multiple identical GPUs. The SPIRIT backbone contains approximately 122.2M parameters. 
Training on the 1,567-instance synthetic training split requires approximately 41 hours for Stage~I and an additional 15 hours for Stage~II. Inference takes approximately 80\,ms per sample. We note that high-resolution rendering is memory-intensive, and GPU memory management becomes important in larger-resolution settings.

\subsection{Evaluation Metrics}
\label{app:metrics}

\paragraph{Signal Error.}
Signal error is computed in the frequency (phase-history) domain as the mean absolute complex residual over valid samples after the common preprocessing described above:
\begin{equation}
\mathcal{L}_{\mathrm{sig}} = \frac{1}{|\Omega|}\sum_{q\in\Omega} \left| S_{\mathrm{pred},q} - S_{\mathrm{gt},q} \right|,
\end{equation}
where $\Omega$ indexes the valid phase-history entries used for evaluation. For image-domain baselines, the predicted image is first passed through the same forward operator before comparison.

\paragraph{Image Metrics.}
SSIM and PSNR are computed on 2D magnitude images obtained via standard backprojection:
\begin{equation}
I_{\mathrm{pred}} = \mathrm{BP}(S_{\mathrm{pred}}), \quad
I_{\mathrm{gt}} = \mathrm{BP}(S_{\mathrm{gt}}).
\end{equation}
SSIM is computed with data range normalized to $[0,1]$.
PSNR is defined as:
\begin{equation}
\mathrm{PSNR} = 20 \log_{10} \left( \frac{1}{\sqrt{\mathrm{MSE}}} \right).
\end{equation}

Image-level metrics depend on the chosen image-formation operator and therefore complement rather than replace geometry- and signal-domain evaluation. The same metric definitions are used for the held-out MITO evaluation, with its common registered reference geometry and measured coherent data.

\paragraph{Chamfer Distance.}
Chamfer Distance is computed from metric scene coordinates and reported in millimeters throughout the paper. Synthetic coordinates stored in meters are converted to millimeters for reporting; MITO geometry is evaluated in its registered metric coordinate frame.
For SPIRIT, predicted surface vertices are directly compared with the reference point set. For outputs represented on a reconstruction grid, a common fixed threshold and Marching Cubes surface extraction are used before geometric evaluation. CD is computed as the symmetric mean nearest-neighbor distance between the two point sets. No method-specific post-hoc alignment is applied in the controlled benchmark.

\subsection{Reproducibility Notes}
Data partitions and deterministic evaluation masks use random seed 42. The main SPIRIT comparison reports mean $\pm$ standard deviation over three independent training runs, whereas matched controls use the fixed schedules described above. All methods share the same evaluation pipeline and post-processing steps.
Runtime comparisons are not emphasized, as this work focuses on reconstruction fidelity under extreme sparsity rather than real-time performance.

\section{Extended Experiments}
\label{app:ablation}

This appendix provides a comprehensive experimental analysis of SPIRIT beyond the main quantitative results.
We focus on understanding performance under practical sensing conditions and isolating the effect of key design choices.
Specifically, we examine (i) the interaction between Stage~1 geometry learning and Stage~2 refinement,
(ii) the effect of aperture coverage on SAR image formation,
(iii) held-out evaluation on real MITO measurements, 
(iv) robustness to measurement noise,
and
(v) the role of constrained signal refinement.
Together, these studies characterize when the proposed design is effective, how its components interact, and where the evidence remains limited.

% ------------------------------------------------------------
% ------------------------------------------------------------
\subsection{Two-Stage Ablation: Geometry Learning and Signal Refinement}

We examine the interaction between Stage~1 geometry learning and Stage~2 signal-driven refinement under extreme sparsity.
Table~\ref{tab:ablation_core} summarizes the main ablation results.

Table~\ref{tab:ablation_core} in the main paper reports the complete stage-wise comparison. The qualitative examples below visualize the geometric drift observed for unconstrained Stage~II optimization.

\begin{figure*}[ht] \centering \setlength{\tabcolsep}{2pt} \begin{tabular}{ccc} \textbf{Snacks} & \textbf{Dish Towel} & \textbf{Shoes} \\ \includegraphics[width=0.25\linewidth]{spirit_content/figure/target_example/Black_Forest_Fruit_Snacks_28_Pack_Grape_090/pointcloud_gt.pdf} & \includegraphics[width=0.25\linewidth]{spirit_content/figure/target_example/Cole_Hardware_Dishtowel_Blue_147/pointcloud_gt.pdf} & \includegraphics[width=0.25\linewidth]{spirit_content/figure/target_example/Court_Attitude_170/pointcloud_gt.pdf} \\ \includegraphics[width=0.25\linewidth]{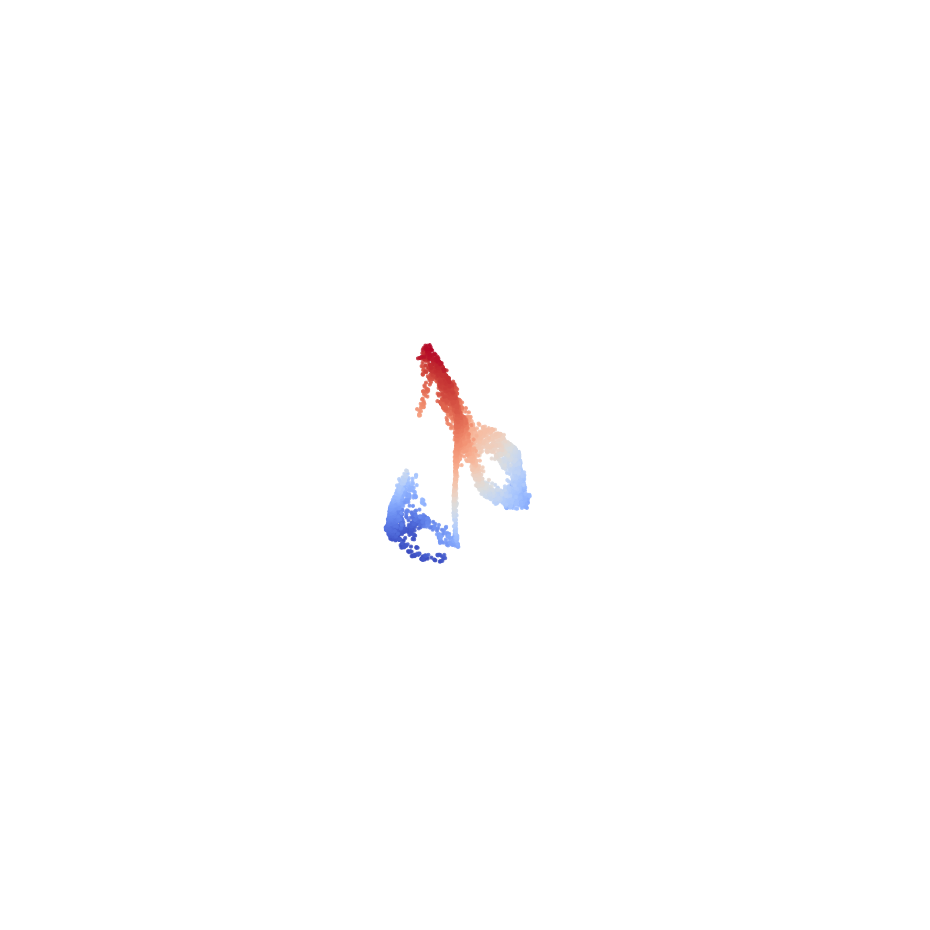} & \includegraphics[width=0.25\linewidth]{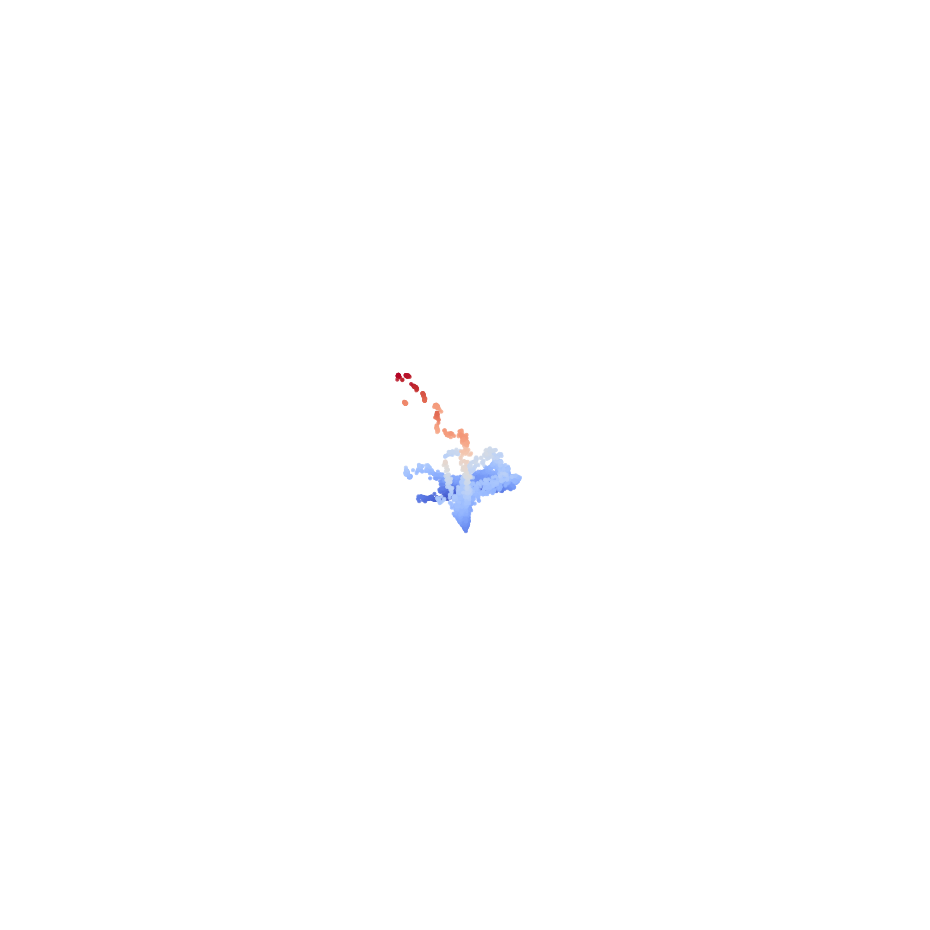} & \includegraphics[width=0.25\linewidth]{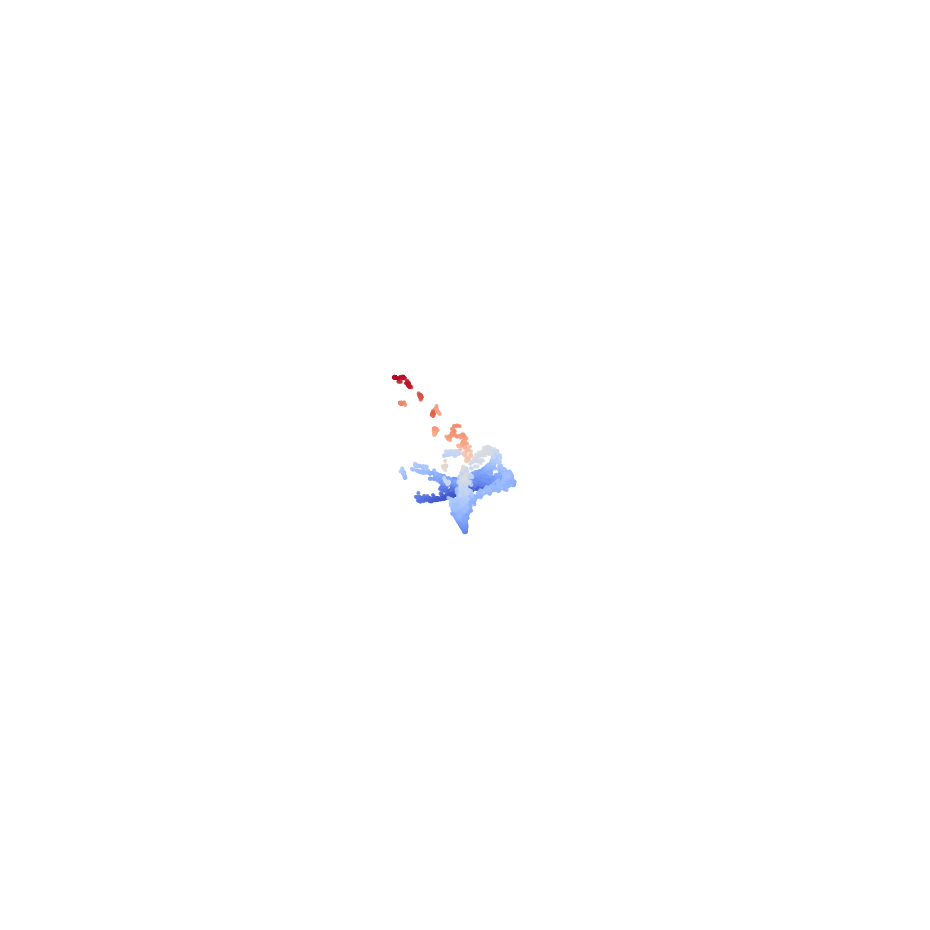} \\ \end{tabular} \vspace{-4pt} \caption{Geometry collapse under unconstrained Stage~2 fine-tuning. The top row shows ground truth, while the bottom row shows Stage~2 full fine-tuning without geometric constraints. Despite improved image appearance, unconstrained optimization leads to geometric drift and collapse.} \label{fig:stage2_collapse} \vspace{-6pt} \end{figure*}

Unconstrained refinement improves image-level metrics but can induce geometric drift,
as illustrated in Fig.~\ref{fig:stage2_collapse},
motivating the constrained refinement strategy adopted in Stage~2.

% ------------------------------------------------------------
% ------------------------------------------------------------
\subsection{Aperture Coverage and SAR Image Formation}

\begin{figure*}[ht] \centering \setlength{\tabcolsep}{1.5pt} \begin{tabular}{cccccccc} \textbf{Snacks} & \textbf{Dishtowel} & \textbf{Shoes} & \textbf{Hair Kit} & \textbf{Pencil Case} & \textbf{Card} & \textbf{Flats} & \textbf{Marker} \\ \includegraphics[width=0.115\linewidth]{spirit_content/figure/target_example/Black_Forest_Fruit_Snacks_28_Pack_Grape_090/pred.pdf} & \includegraphics[width=0.115\linewidth]{spirit_content/figure/target_example/Cole_Hardware_Dishtowel_Blue_147/pred.pdf} & \includegraphics[width=0.115\linewidth]{spirit_content/figure/target_example/Court_Attitude_170/pred.pdf} & \includegraphics[width=0.115\linewidth]{spirit_content/figure/target_example/Just_For_Men_Mustache_Beard_Brushin_Hair_Color_Gel_Kit_Jet_Black_M60_375/pred.pdf} & \includegraphics[width=0.115\linewidth]{spirit_content/figure/target_example/Olive_Kids_Butterfly_Garden_Pencil_Case_577/pred.pdf} & \includegraphics[width=0.115\linewidth]{spirit_content/figure/target_example/pokemon_yellow_669/pred.pdf} & \includegraphics[width=0.115\linewidth]{spirit_content/figure/target_example/White_Rose_Tieks_Leather_Ballet_Flats_with_Floral_Rosettes_970/pred.pdf} & \includegraphics[width=0.115\linewidth]{spirit_content/figure/target_example/small_marker_041/pred.pdf} \\ \includegraphics[width=0.115\linewidth]{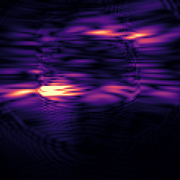} & \includegraphics[width=0.115\linewidth]{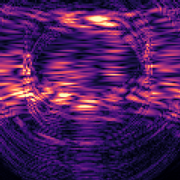} & \includegraphics[width=0.115\linewidth]{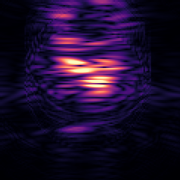} & \includegraphics[width=0.115\linewidth]{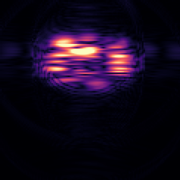} & \includegraphics[width=0.115\linewidth]{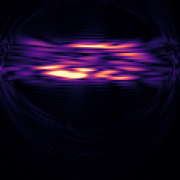} & \includegraphics[width=0.115\linewidth]{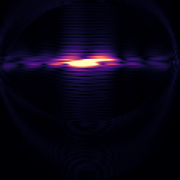} & \includegraphics[width=0.115\linewidth]{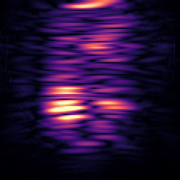} & \includegraphics[width=0.115\linewidth]{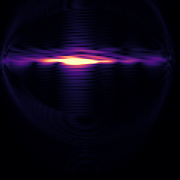} \\ \includegraphics[width=0.115\linewidth]{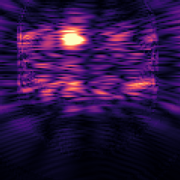} & \includegraphics[width=0.115\linewidth]{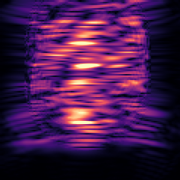} & \includegraphics[width=0.115\linewidth]{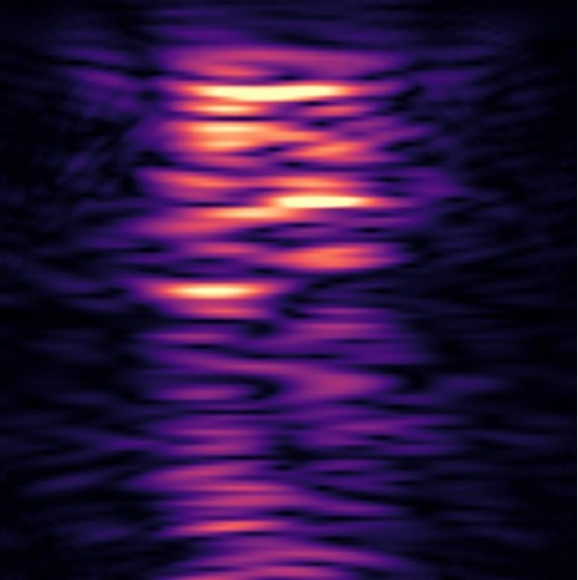} & \includegraphics[width=0.115\linewidth]{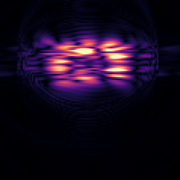} & \includegraphics[width=0.115\linewidth]{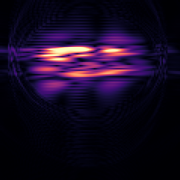} & \includegraphics[width=0.115\linewidth]{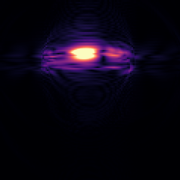} & \includegraphics[width=0.115\linewidth]{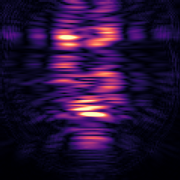} & \includegraphics[width=0.115\linewidth]{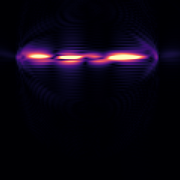} \\ \includegraphics[width=0.115\linewidth]{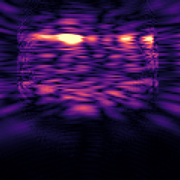} & \includegraphics[width=0.115\linewidth]{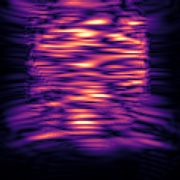} & \includegraphics[width=0.115\linewidth]{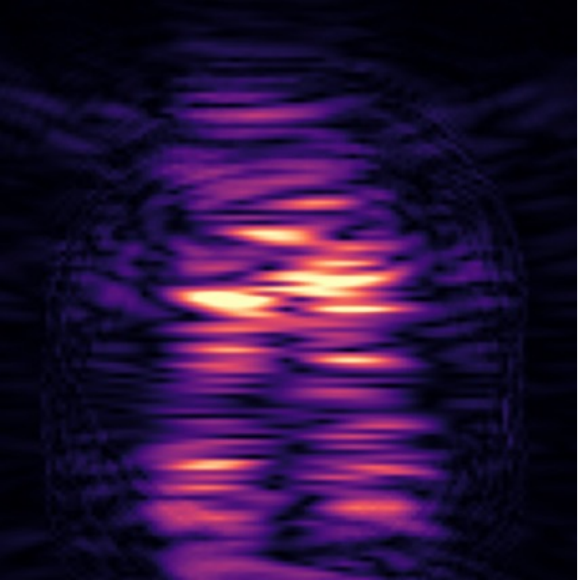} & \includegraphics[width=0.115\linewidth]{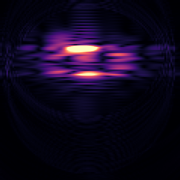} & \includegraphics[width=0.115\linewidth]{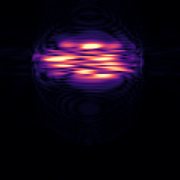} & \includegraphics[width=0.115\linewidth]{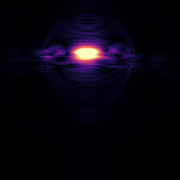} & \includegraphics[width=0.115\linewidth]{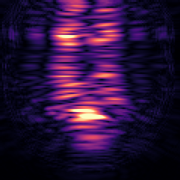} & \includegraphics[width=0.115\linewidth]{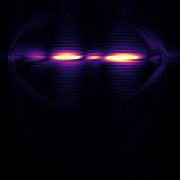} \\ \end{tabular} \vspace{-4pt} \caption{Reconstruction under increasing aperture coverage. Rows correspond to 1-, 3-, and 6-scan inputs. Additional scanlines provide complementary angular and spatial-frequency measurements, reducing ambiguity relative to the 1-scan setting. Sparse sampling nevertheless leaves unobserved support and does not uniquely determine the scene.} \label{fig:sar_scan_comparison} \vspace{-6pt} \end{figure*}

Figure~\ref{fig:sar_scan_comparison} visualizes the effect of retaining different amounts of aperture support. A single scanline provides only a restricted subset of the spatial-frequency and angular observations available in the full aperture. Adding scanlines introduces additional viewpoint-dependent measurements and reduces, but does not eliminate, the ambiguity of the inverse problem. The experiment therefore illustrates complementarity among retained aperture positions; it should not be interpreted as evidence that sparse sampling preserves all scene information.

\subsection{Real-World Evaluation on MITO}
\label{sec:real_world}

We evaluate the final SPIRIT system on held-out physical measurements from the MITO dataset~\cite{dodds2025mito}. The real-data split is object-disjoint: 40 physical objects are used to adapt the Stage~II residual-refinement branch, 10 for validation, and 10 for testing; the two $90^\circ$-separated captures of each object remain in the same partition. Stage~I geometry learning uses only synthetic measurements, and no real test capture is used during residual adaptation. Because the full pipeline has access to a disjoint real training partition, we describe this as synthetic geometry learning plus real residual adaptation rather than zero-shot synthetic-to-real transfer.

The real measurements contain hardware response, trajectory/calibration residuals, material-dependent scattering, occlusion, and multipath effects that are not reproduced exactly by either the synthetic generator or SPIRIT's single-scattering renderer. For geometric evaluation, we use the dataset-provided reference meshes registered to the radar coordinate frame using the released calibration and object pose. The same reference is used for every method and no method-specific post-hoc alignment is applied. We report Chamfer Distance in millimeters, phase-history $\ell_1$ error, and PSNR/SSIM on backprojected magnitude images using the same metric definitions as the controlled benchmark.

\begin{table}[ht]
\centering
\caption{Quantitative comparison on the object-disjoint MITO real-world test split after residual-only Stage~II adaptation on the disjoint real training split. The listed methods are matched controls under the common evaluation pipeline rather than claimed faithful reproductions of every recent radar system. CD is computed against the common registered reference geometry; signal and image metrics use the corresponding measured/reference radar products. Lower is better for CD and signal error; higher is better for PSNR/SSIM.}
\label{tab:real_world}
\small
\setlength{\tabcolsep}{4.5pt}
\begin{tabular}{lcccc}
\toprule
Method & SSIM $\uparrow$ & PSNR (dB) $\uparrow$ & CD (mm) $\downarrow$ & Signal $\ell_1$ $\downarrow$ \\
\midrule
Backprojection & 0.452 & 18.15 & 18.52 & 0.851 \\
Compressed Sensing & 0.484 & 10.42 & 6.15 & 0.621 \\
Implicit Neural Field & 0.621 & 12.10 & 4.85 & 0.554 \\
UNet (Image Domain) & 0.655 & 13.50 & 4.20 & 0.482 \\
Volumetric Neural Field & 0.710 & 14.85 & 6.50 & \textbf{0.353} \\
Direct Volumetric Field & 0.685 & 10.10 & 5.85 & 0.401 \\
\midrule
\textbf{SPIRIT (Ours)} & \textbf{0.805} & \textbf{20.45} & \textbf{3.15} & 0.362 \\
\bottomrule
\end{tabular}
\end{table}

SPIRIT obtains the lowest Chamfer Distance and the highest PSNR/SSIM among the evaluated methods on the held-out physical test objects. Its signal error is competitive but not the lowest: the volumetric neural-field control reaches $0.353$ versus SPIRIT's $0.362$. This mirrors the controlled ablations and reinforces the distinction between signal residual and geometric correctness. Because the experiment remains a controlled, object-scale near-field setting, we do not extrapolate these results to cluttered, multi-object, airborne, or strongly multipath SAR scenes.

% ------------------------------------------------------------
\paragraph{Auxiliary-sweep scope.}
The following robustness and regularization sweeps are auxiliary controlled analyses used to study within-table trends. Their absolute CD values are not used to support the primary benchmark claims in Table~\ref{tab:quantitative_main}; those claims are based on the object-disjoint benchmark and MITO protocols stated explicitly above.

\subsection{Noise Robustness}

To assess robustness to measurement noise, we inject additive complex Gaussian noise at different signal-to-noise ratios (SNRs).
For each noise level, we report Chamfer Distance (mm) and signal $\ell_1$ error.
Results are summarized in Table~\ref{tab:noise}.
\begin{table}[ht]
\centering
\caption{
Noise robustness analysis under varying SNR levels.
Each entry reports Chamfer Distance (mm) / Signal $\ell_1$ error.
Lower values indicate better geometric fidelity and signal consistency.
}
\label{tab:noise}
\begin{tabular}{lcccc}
\toprule
Method & $\infty$ dB & 30 dB & 20 dB & 10 dB (High Noise) \\
\midrule
Backprojection 
& 16.8 / 0.61 & 22.5 / 0.79 & 26.4 / 0.78 & 55.3 / 0.82 \\

Compressed Sensing 
& 8.9 / 0.24 & 12.1 / 0.43 & 17.6 / 0.52 & 36.2 / 0.72 \\

Implicit Neural Field 
& 5.6 / 0.32 & 8.1 / \textbf{0.06} & 9.4 / \textbf{0.11} & 19.7 / 0.86 \\

UNet (Image Domain) 
& 4.1 / 0.64 & 5.0 / 0.72 & 8.8 / 0.87 & 15.6 / 0.91 \\

Volumetric Neural Field 
& 7.4 / 0.37 & 15.2 / 0.45 & 19.6 / 0.51 & 48.9 / 0.74 \\

Direct Volumetric Field 
& 4.0 / 0.40 & 5.8 / 0.71 & 10.5 / 0.13 & 22.4 / 0.58 \\

\midrule
\textbf{SPIRIT (Ours)} 
& \textbf{2.9 / 0.20} 
& \textbf{3.2} / 0.28 
& \textbf{4.6} / 0.48 
& \textbf{7.9 / 0.57} \\

\bottomrule
\end{tabular}
\end{table}

SPIRIT has the lowest Chamfer Distance at every tested noise level. Signal error is not uniformly lowest, again showing that signal residual and geometric fidelity need not rank methods identically.

% ------------------------------------------------------------
\subsection{Effect of Identity Regularization}

We vary the identity-regularization coefficient $\lambda_{\mathrm{id}}$ while keeping the residual-refiner architecture fixed. Table~\ref{tab:beta} reports the resulting geometry--signal trade-off.

\begin{table}[ht]
\centering
\caption{Sensitivity to the identity-regularization coefficient $\lambda_{\mathrm{id}}$.}
\label{tab:beta}
\begin{tabular}{lcc}
\toprule
$\lambda_{\mathrm{id}}$ & CD (mm) $\downarrow$ & Signal $\ell_1$ $\downarrow$ \\
\midrule
0.0 (No identity reg.) & 4.8 & \textbf{0.22} \\
0.01 & 3.5 & 0.24 \\
0.05 & \textbf{2.1} & 0.28 \\
0.1 & 2.8 & 0.35 \\
0.5 & 5.5 & 0.52 \\
\midrule
No Refiner & 6.2 & 0.58 \\
\bottomrule
\end{tabular}
\end{table}

The sweep exposes a Pareto trade-off rather than a single metric optimum: weak identity regularization gives the smallest signal residual, while moderate regularization gives the lowest geometric error. This behavior reinforces our decision to interpret signal and geometry metrics jointly.

\section{Discussion, Limitations, and Societal Impact}
\label{app:discussion}

This section discusses the applicability, limitations, and broader implications of SPIRIT.
Rather than repeating experimental results, we aim to clarify when the method is effective, where it may fail, and how its design choices shape these behaviors.

\subsection{Applicability and Failure Modes}

\paragraph{Scope of empirical evidence.}
Our primary mechanism and attribution conclusions are established on the controlled synthetic benchmark, where the generating geometry and measurement process are fully specified. Appendix~\ref{sec:real_world} additionally evaluates held-out physical MITO measurements after residual-only adaptation on a disjoint real training split. These measurements provide evidence under model--reality mismatch, but the experiment remains controlled, object-scale, and near-field; it does not establish zero-shot generalization or performance in operational multi-object or remote-sensing SAR scenes.

\paragraph{Single-Object Assumption.}
SPIRIT is primarily designed for scenes dominated by a single main object.
The controlled geometry experiments in this work use centered, isolated objects. The current structured $N\times N$ surface grid imposes a strong inductive bias toward a single coherent surface and is not designed to robustly represent arbitrary numbers of disconnected components or volumetric scatterers. We therefore restrict the geometric conclusions to this single-object regime; multi-object reconstruction requires a more flexible scene representation and dedicated evaluation. In particular, the present experiments do not cover clutter, layover, ground-object double bounce, vegetation volume scattering, or operational airborne/spaceborne SAR geometries.

% \paragraph{Multi-Object Scenarios.}
% Failures in multi-object settings arise from a combination of implementation bias and fundamental ambiguity.
% From an implementation perspective, the single-object training regime induces a strong prior toward connected shapes, which could be addressed in future work through instance-level supervision or multi-surface representations.
% From an information-theoretic perspective, severe occlusion and inter-object multipath under sparse SAR collection introduce genuine ambiguity.
% Without additional viewpoints or increased bandwidth to provide new lines of sight or time-of-flight resolution, separating multiple objects becomes an ill-posed inverse problem.
% In such cases, data-driven priors cannot recover information that is physically absent due to shadowing.

\subsection{Attribution Scope}
The main experiments separate two sources of improvement only to the extent supported by the available controls: Table~\ref{tab:quantitative_main} compares matched representation/reconstruction controls, while Table~\ref{tab:ablation_core} isolates the Stage~I/Stage~II optimization choices within SPIRIT. This is not a fully factorial comparison in which every representation is trained with identical parameter count, optimizer, and staged schedule. We therefore attribute the reported performance to the combined geometry-first system and claim only that the structured surface prior is effective in the studied regime, not that fixed topology is uniquely necessary or universally optimal.

\subsection{Limitations of Structured Surface Representation}

\paragraph{Surface Parameterization and Self-Intersection.}
The structured surface representation used in SPIRIT does not enforce topological constraints.
Each point in the $N \times N$ grid predicts its $(x, y, z)$ coordinates independently, allowing self-intersection and folding.
This flexibility enables the representation of complex shapes, such as concavities or overlapping structures, but does not guarantee manifold consistency.

\paragraph{Holes and Thin Structures.}
Failures on holes and thin structures arise from both measurement ambiguity and representation limits. Sparse aperture coverage may provide insufficient support for weak, shadowed, or self-occluded structures, in which case reconstruction is necessarily prior-dependent. The fixed-resolution surface grid additionally limits features below its point spacing and favors coherent topology. Increasing aperture density can improve conditioning, while wider or more diverse trajectories may introduce new lines of sight; neither change removes the representation limits of the current fixed-topology surface. A visually plausible completion in a poorly observed region should therefore not be interpreted as evidence that the corresponding geometry was uniquely determined by the measurement.

\paragraph{Future Directions.}
Addressing multi-object scenes may require multi-grid or instance-wise surface representations combined with explicit segmentation.
Recovering thin structures or complex topology may benefit from hybrid explicit--implicit surface formulations or adaptive resolution strategies.
These extensions are left for future work.

\subsection{Physics-Constrained Modeling}

\paragraph{Role of the Signal Refiner.}
The refiner is a statistical model-mismatch compensator rather than an electromagnetic simulator. Identity regularization and a small residual weight constrain its contribution, but do not guarantee that every correction corresponds to a physically modeled scattering mechanism. Under strong multipath, calibration error, or other unmodeled effects, the refiner may absorb signal structure that is not explained by the recovered surface. For this reason, improvements in signal error are interpreted jointly with geometric accuracy rather than as independent evidence of geometry recovery.

\subsection{Risks and Trade-offs in Signal Refinement}
\paragraph{Geometry Collapse.}
Unconstrained signal-driven optimization can lead to geometric collapse, in which the reconstructed surface deforms to minimize signal loss rather than accurately representing the actual structure.
Our ablation studies show that this behavior is configuration-dependent but poses a high risk without proper constraints.

% \paragraph{Mitigation and Pareto Frontier.}
% Collapse is effectively mitigated by (i) maintaining intense geometric supervision anchored to Stage~1, (ii) assigning relatively low weight to signal loss, and (iii) restricting gradient flow through the refiner.
% Due to mismatches between the approximate renderer and real-world physics, the geometry that best matches the signal may not be the actual geometry.
% This induces a geometry--signal Pareto frontier, and SPIRIT aims to operate at a balanced point where geometry remains physically plausible while signal consistency is improved.

% \subsection{Runtime \& Memory}
% SPIRIT is designed for offline reconstruction under extreme aperture sparsity, prioritizing geometric fidelity and physical consistency over computational efficiency. As a result, we do not explicitly optimize runtime or memory usage, and our evaluation does not include a detailed comparison of computational cost across methods. In addition, while the signal refinement module is constrained through identity regularization and gradient blocking, we do not impose an explicit frequency-domain or spectral constraint; analyzing the learned frequency response of the refiner and establishing tighter bounds on its effect remain important directions for future work. Finally, our forward model adopts a single-scattering approximation and soft visibility, which may limit performance in scenes with strong multipath or complex material responses.

\section{Potential Applications and Broader Scope}

The sparse-aperture inverse-rendering formulation may be relevant to object-scale millimeter-wave applications in which dense spatial scanning is costly, including security sensing, human-centric radar perception, and close-range material or agricultural sensing. These domains share a collection-efficiency challenge, but they also introduce scene statistics and scattering effects not evaluated in this work \cite{Adonis25Liu,hydra24liu, Proteus25Liu, liu2025hydrabench, liu26aura, gan2023poster}.

SPIRIT has not been validated here for concealed-object detection, face authentication, vegetation reconstruction, airborne SAR, or other downstream tasks. The current evidence combines controlled synthetic attribution with held-out physical-object MITO evaluation; application-specific deployment would require dedicated measurements, forward-model validation, and task-level evaluation.

\section{AI Use Statement}
\label{app:ai-statement}
We used generative AI tools, including OpenAI ChatGPT, OpenAI Codex, and Google Gemini, in several stages of this work.

For writing support, these tools were used to edit and polish author-provided text, with the goal of improving clarity, readability, organization, grammar, and presentation without altering the intended scientific meaning.

For retrieval and discovery, generative AI was used to assist in identifying potentially relevant literature, locating related research directions, and organizing references for the related-work discussion. All cited references were independently checked by the authors against their original sources before inclusion.

For research ideation and execution, generative AI was used to help structure research questions and technical ideas, examine alternative system designs, provide feedback on experimental plans, and assist with code development, debugging, and implementation. The authors determined the research objectives, methodological choices, experimental protocols, and final system design.

Generative AI was also used to help draft and revise portions of the manuscript based on outlines, technical descriptions, experimental results, and scientific content provided and verified by the authors. All generated or revised text was reviewed, edited, and approved by the authors before inclusion in the final manuscript.

Generative AI was not used to fabricate experimental data, numerical results, or evaluation outcomes. All reported results were produced by executing the implemented experimental pipeline on the datasets described in the paper. AI-assisted code, experimental configurations, citations, technical statements, results, interpretations, and scientific claims were inspected and verified by the authors.

The authors retain full responsibility for the research design, implementation, analysis, interpretation, claims, conclusions, and final content of the paper.

\end{document}